\documentclass{article}

\usepackage[eandd, preprint, nonatbib]{neurips_2026}

\usepackage[utf8]{inputenc} 
\usepackage[T1]{fontenc}    
\usepackage{hyperref}       
\usepackage{url}            
\usepackage{booktabs}       
\usepackage{amsfonts}       
\usepackage{nicefrac}       
\usepackage{microtype}      
\usepackage[table,xcdraw]{xcolor}         
\usepackage{graphicx}
\usepackage{wrapfig}
\usepackage{enumitem}
\usepackage[mode = text, range-phrase = {-}]{siunitx}

\usepackage[natbib,style=numeric,sorting=none]{biblatex}
\title{Static Recovery Is Not Dynamic Stability: Dynamics-Aware Benchmarking of Protein Motif Scaffolding}

\author{%
  Emile de Bruyn$^{1*}$ \\
  \And
  Moritz Schäffler$^{1*}$ \\
  \And
  Jannik Schneider$^{1*}$ \\
  \And
  Sebastian R. Schmidt$^{1}$\\
  \And
  Chanwoong Hwang$^{2}$ \\
  \And
  Aleena Siji$^{3}$ \\
  \And
  Utkarsh Upadhyay$^{1}$ \\
  \And
  Stefan Bauer$^{3}$ \\
  \And
  Alexander Schug$^{4}$ \\
  \And
  Stefan Kesselheim$^{1,8}$ \\
  \And
  Artur Yakimovich$^{2,5,6,7}$ \\
  \And
  Alina Bazarova$^{1}$ \\
  \And
 {\normalfont $^*$Contributed equally} \quad
 {\normalfont $^1$Jülich Supercomputing Centre, Forschungszentrum Jülich} \\
 {\normalfont $^2$Center for Advanced Systems Understanding (CASUS), Görlitz, Germany} \\
 {\normalfont $^3$Helmholtz Munich} \quad
 {\normalfont $^4$Scientific Computing Center, Karlsruhe Institute of Technology} \\
 {\normalfont $^5$Helmholtz-Zentrum Dresden-Rossendorf e. V. (HZDR), Dresden, Germany} \\
 {\normalfont $^6$Institute of Computer Science, University of Wrocław, Wrocław, Poland} \\
 {\normalfont $^7$Cluster of Excellence Physics of Life, TU Dresden, Dresden, Germany} \\
 {\normalfont $^8$Department of Physics, University of Cologne} \\
 \texttt{\{e.de.bruyn,mo.schaeffler,ja.schneider,s.schmidt,u.upadhyay,}\\
 \texttt{s.kesselheim,a.bazarova\}@fz-juelich.de}\\
 \texttt{\{aleena.siji,stefan.bauer\}@helmholtz-munich.de}\\
 \texttt{hcw1027@gmail.com} \quad \texttt{alexander.schug@kit.edu} \quad \texttt{a.yakimovich@hzdr.de}
}

\begin{document}
\begin{refsection}

\maketitle

\begin{abstract}
Protein motif scaffolding benchmarks aim to measure whether generative models can build scaffolds around functionally important residue sets. Their conclusions depend on two coupled choices: which motifs are tested and which criteria define success. Existing benchmarks typically use small hand-curated motif sets and static refolding-based criteria such as motif recovery, self-consistency, and structural novelty. However, protein function often depends on conformational ensembles and transitions, raising the question of whether static refolding success translates to dynamic fidelity of motifs and scaffolds. Here, we address both benchmark scope and success criteria. We introduce BILBY-PRO, a systematically derived benchmark of structurally conserved functional motifs from PROSITE-linked experimental structures, comprising 220 cases from 174 PROSITE motif-pattern entries of varying conformations. We evaluate six recent scaffold-generation models and find that performance separates into target coverage and unique-solution yield: RFDiffusion and Proteina solve more targets, whereas ESM3 produces more unique successful scaffolds on the targets they solve. We further introduce BILBY-MD, a reference-based all-atom molecular dynamics evaluation framework, which we apply to statically solved designs from the original MotifBench benchmark to compare motif drift, scaffold drift, motif-contact preservation, and motif-dihedral free-energy landscapes against reference simulations. Of 55 statically solved model–problem pairs, 31 satisfy our MD-supported criterion and only 8 pass all checks. Refolding success is therefore useful evidence for local motif recovery, but not sufficient evidence for dynamical fidelity of the full scaffold, exposing misalignment between static benchmark success and downstream dynamical behavior. The mismatch is motif-dependent: compact, self-contained motifs are often MD-supported, whereas interface-like or context-dependent motifs fail more often. We show that motif-scaffolding models have varying performance on three complementary axes: target coverage, unique-solution yield, and MD-supported dynamical fidelity.
\end{abstract}

\section{Introduction}

Protein motif scaffolding asks whether a generative model can build a foldable protein scaffold around a prescribed set of functionally important residues. This task is central to the emerging field of generative protein design \cite{tischer10design,wang2022scaffolding}, making benchmark quality especially important. A useful benchmark should reveal not only whether a model can reproduce a static motif geometry, but also which functional motifs are scaffoldable, which models fail systematically, and whether nominally successful designs remain stable under physically realistic conformational fluctuations \cite{Cho2025}.

Existing motif-scaffolding benchmarks rely on static structure-level metrics: they evaluate a single predicted protein structure rather than its conformational behaviour over time. Standard evaluations ask whether a generated design can be refolded to recover the target motif in this single structure and whether the resulting scaffold is novel with respect to known proteins. While these are sensible and computationally scalable criteria, they do not assess whether the recovered motif remains stable under thermal fluctuations. Consequently, designs that satisfy conventional static benchmark criteria may nevertheless lose the required motif geometry during their natural conformational dynamics, potentially compromising their functional plausibility. This distinction is particularly important for catalytic and binding-site motifs, whose function depends on maintaining the correct geometry throughout their conformational dynamics rather than in a single predicted structure~\cite{Warshel2006,GRUNBERG20042125,Akdel2022}.

A central challenge in benchmarking protein scaffold generation is that evaluation protocols implicitly define what is considered a “successful” design, by the choice of metrics used. Here, we focus on the MotifBench~\cite{motifbench} benchmark as a representative case study for scaffold-design evaluation. MotifBench adopts a now widely used evaluation paradigm that emerged from earlier generative modeling efforts such as RFDiffusion~\cite{rfdiffusion}, FrameFlow~\cite{frameflow}, and Genie2~\cite{genie2}, and is increasingly reused in broader benchmarks including ProteinBench~\cite{proteinbench} and GeomMotif~\cite{geommotif}. This paradigm is based on an inverse-folding plus refolding oracle, where candidate backbones are sequence-designed using ProteinMPNN~\cite{proteinMPNN} and subsequently validated via single-sequence structure prediction with ESMFold~\cite{ESMFold} (or, in earlier work, AlphaFold2~\cite{AF2}). MotifBench develops this into a standardized and reproducible benchmark with fixed sampling protocols, success thresholds, and aggregation metrics, thereby reflecting its emergence as a de facto community baseline.

Here, we revisit the validity of this evaluation paradigm by assessing the dynamical stability of MotifBench ``solved'' designs. To cover the variety of state-of-the-art generative models for protein design and build upon previous benchmarks~\cite{motifbench}, we chose diffusion models, namely RFDiffusion~\cite{rfdiffusion} and Genie2~\cite{genie2}, the flow-matching models FrameFlow~\cite{frameflow} and Proteina~\cite{proteina} and for the Protein Language Model domain DPLM2~\cite{dplm2} and ESM3~\cite{esm3}. In this study, molecular dynamics (MD) analysis across diverse generative models indicates that many nominally ``solved'' cases exhibit substantial scaffold instabilities, with only a minority remaining fully stable under dynamical evaluation. These results suggest that the ProteinMPNN+ESMFold pipeline primarily captures self-consistency under a single folding model and tends to emphasize local motif reconstruction over global structural stability. 

Another limitation is dataset scale and construction. Although the original MotifBench benchmark has proven useful, it contains only 30 manually curated test cases. Such small hand-picked collections can be highly informative, yet they inevitably entangle method performance with dataset composition. A larger and more systematically derived benchmark is therefore valuable even if it preserves the original ranking structure, because it reduces dependence on a narrow set of canonical cases and increases statistical resolution. 

In this work, we introduce BILBY (Benchmarking Interactions, Landscapes, Backbone fidelity, and dYnamics), comprising two complementary resources for functional motif-scaffolding evaluation. First, BILBY-PRO is a systematically constructed benchmark of 220 functional motif-scaffolding problems derived from 174 PROSITE motif-pattern entries linked to experimentally resolved structures. Constructed independently of the original MotifBench dataset, BILBY-PRO adopts a compatible task representation and evaluation protocol. Second, we evaluate six recent scaffold-generation methods on both the original MotifBench panel and BILBY-PRO, revealing model-specific trade-offs between target coverage and unique-solution yield. Third, BILBY-MD provides a reference-based, dynamics-aware evaluation framework using explicit-solvent all-atom MD to assess motif geometry, scaffold stability, contact preservation, and conformational ensemble agreement. Here, we apply BILBY-MD to designs evaluated on the original MotifBench panel. Together, these contributions separate benchmark composition, static design success, and dynamical fidelity, demonstrating when static motif recovery does, and does not, translate into dynamically credible scaffold behavior.

\section{BILBY-PRO: PROSITE-derived Functional Motif Benchmark}

By defining benchmark instances through uniform structural-conservation criteria rather than manual case selection, BILBY-PRO provides a consistent and scalable foundation for evaluating motif-scaffolding methods across a substantially broader range of functional motifs.

PROSITE is a curated database of protein families, domains, and functional sites represented by conserved sequence patterns and profiles~\cite{prosite}. We construct BILBY-PRO from functional motifs identified in PROSITE-linked experimental structures. Rather than selecting scaffolding tasks manually, we start from curated PROSITE sequence patterns, identify their occurrences in experimentally resolved protein structures, and retain only motifs with recurrent three-dimensional conservation across multiple structural instances. This ensures that benchmark cases are defined not only by sequence annotation, but by recurring structural realization of a functional motif.

BILBY-PRO targets \emph{within-chain, structurally conserved functional motif signatures}. Because PROSITE patterns primarily encode linear or near-linear residue signatures, the resulting cases are dominated by compact, contiguous or near-contiguous motifs. Multi-segment cases arise mainly when PROSITE patterns contain long internal wildcard spans; in these cases, matched motif residues are separated in sequence, while intervening residues are treated as scaffold context. We therefore interpret the benchmark as a systematic panel of structurally conserved functional motifs. 

Together, these cases provide a substantially broader collection of structurally conserved functional motif-scaffolding tasks than was previously available while remaining directly compatible with established motif-scaffolding evaluation workflows.

\begin{wraptable}{r}{0.50\linewidth}
\small
\vspace{-2.25em}
\caption{Composition of BILBY-PRO. The benchmark contains compact, structurally conserved within-chain motifs derived from PROSITE-linked experimental structures.}
\vspace{.5em}
\label{tab:dataset_stats}
\centering
\begin{tabular}{lr}
\toprule
Statistic & Count \\
\midrule
Benchmark cases & 220 \\
Unique PROSITE motif-pattern entries & 174 \\
Core assemblies & 179 \\
Ribosomal assemblies & 30 \\
Huge assemblies & 11 \\
Median motif length & 19 residues \\
\bottomrule
\end{tabular}
\vspace{-1.0em}
\end{wraptable}

We derive the test cases by parsing eligible PROSITE entries and their linked Protein Data Bank (PDB, \cite{pdb}) structures, matching sequence patterns to protein chains, mapping accepted matches to residue indices and coordinates, and computing structural conservation across all available instances. We retain candidate motifs with at least five structural instances and median pairwise backbone RMSD below a conservation threshold. We also annotate conformational heterogeneity using Foldseek-derived structural clusters \cite{foldseek}, distinguishing predominantly single-state motifs from \emph{multi-state} motifs that recur in multiple structural configurations. 

BILBY-PRO is exported in a format compatible with standard motif-scaffolding workflows, including motif-only structures, source structures, segment annotations, redesignable positions, and contig specifications, i.e. fixed vs. generative residue blocks that define the scaffold generation task. Table~\ref{tab:dataset_stats} summarizes the benchmark composition. BILBY-PRO contains 220 cases from 174 unique PROSITE motif-pattern entries, because some entries give rise to multiple conformational states, and has a median motif length of 19 residues. 

\section{Static Evaluation of Motif-Scaffolding}
We evaluate recent state-of-the-art generative models for motif scaffolding, covering a variety of different approaches. Most prominently, RFDiffusion \cite{rfdiffusion} is trained on denoising protein structures in the frame representation, starting from the checkpoint of the protein folding model RoseTTAFold \cite{rosettafold}. Genie2 \cite{genie2} is trained similarly but relies on a model backbone derived from the AlphaFold architecture. FrameFlow \cite{frameflow} also reasons over the frame representation however via flow matching. Proteina \cite{proteina} is a scaled flow matching model which utilizes hierarchical fold class labels for conditioning. Lastly, protein language models (PLM) achieved state-of-the-art results and rely on discrete protein representations. In this category we chose DPLM2 \cite{dplm2}, trained with discrete diffusion and ESM3 \cite{esm3}, trained via seq2seq translation of sequence, structure and functional representations.
\newline
For each benchmark case, we assess whether the generated scaffold satisfies the benchmark's refolding-based motif recovery criteria and accompanying novelty filters. We apply this protocol both to the original 30 MotifBench cases and to BILBY-PRO, see Tables \ref{tab:static_leaderboard_motif} and \ref{tab:static_leaderboard_prosite} for the obtained leaderboards.

\subsection{Results}

\begin{figure}[t]
    \hspace{-1.4cm}
    \includegraphics[width=1.15\linewidth]{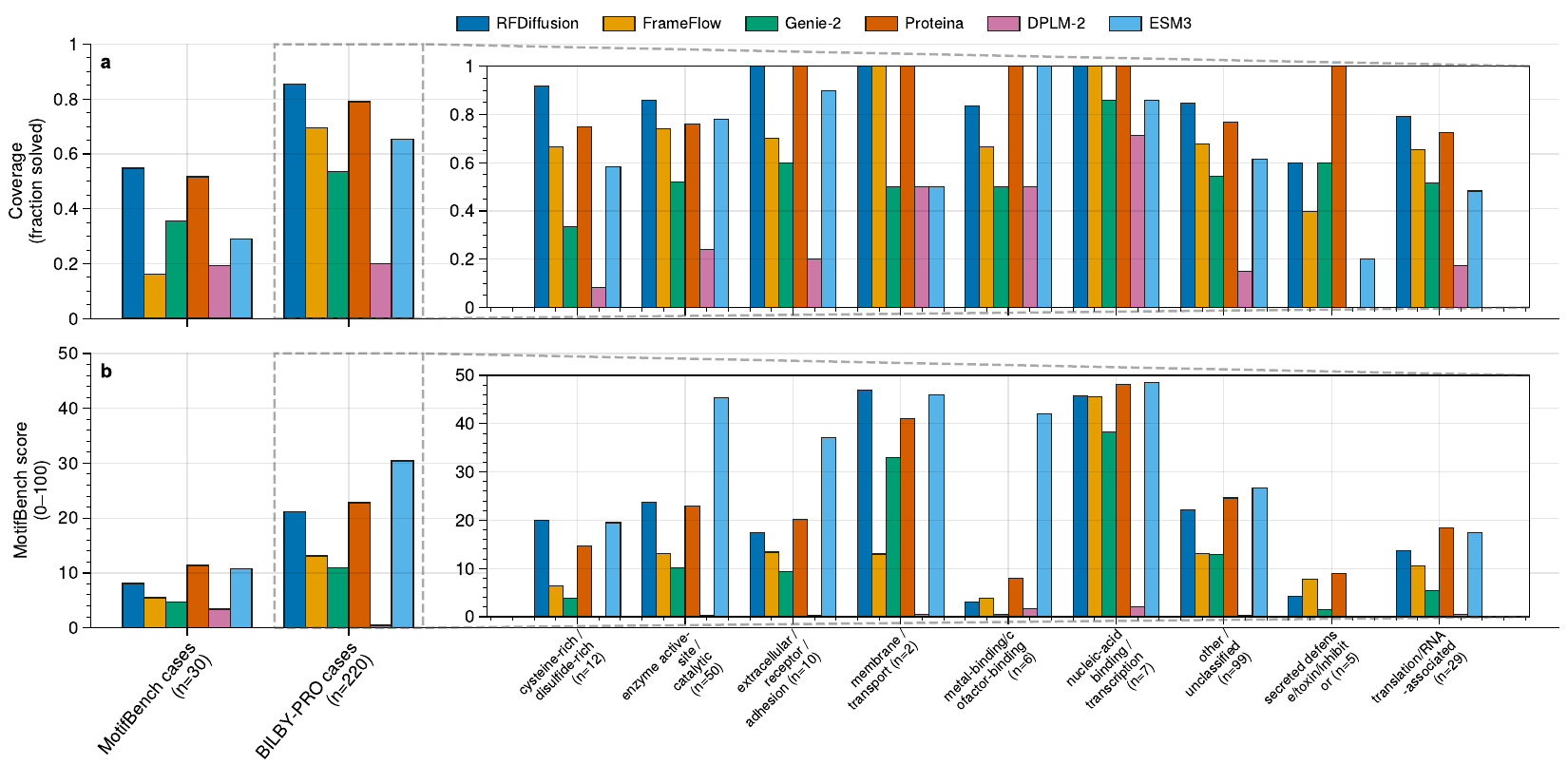}
    \caption{
    {\bf Static motif-scaffolding performance across benchmark sets and broad functional classes.} 
    (a) Coverage is the fraction of cases with at least one successful scaffold among 100 generation attempts. 
    (b) MotifBench score rewards unique successful solutions per case with diminishing return, so additional successes contribute less once a case is already well solved. 
    \newline
    Reporting both metrics separates target-level reliability from unique-solution yield: RFDiffusion and Proteina tend to solve more targets, while ESM3 often achieves higher MotifBench scores on the cases it solves. 
    Numbers in parentheses indicate the number of cases in each subset.
    }
    \label{fig:static_coverage_score}
\end{figure}

 Static motif-scaffolding performance is not one-dimensional (Figure~\ref{fig:static_coverage_score}). Coverage and MotifBench score capture different aspects of model behavior: the coverage measures whether a model solves a target at least once out of the 100 designs, whereas the MotifBench score rewards multiple unique successful solutions under the same computational sampling budget. This distinction is important because BILBY-PRO does not simply make the task uniformly harder or easier, but instead changes the difficulty profile and exposes model-specific coverage--yield trade-offs.

The original 30 MotifBench cases already reveal the trends seen at larger scale. RFDiffusion and Proteina perform best overall, with Proteina producing the most successful solutions. FrameFlow and Genie-2 form an intermediate tier, whereas DPLM-2 remains comparatively weak. ESM3 shows a distinct profile, performing strongly on a subset of motifs where other models often fail despite lower overall success rates. Models also tend to succeed and fail on the same motifs, indicating broadly shared notions of motif difficulty.

Successful-solution counts are strongly correlated among RFDiffusion, Proteina, FrameFlow, Genie-2, and DPLM-2 ($r \approx \numrange{0.61}{0.96}$), with especially high agreement between RFDiffusion and Proteina ($r = \num{0.96}$). FrameFlow also correlates strongly with RFDiffusion ($r = \num{0.85}$) and Genie-2 ($r = \num{0.87}$). In contrast, ESM3 shows weak correlations with all other models ($r \approx \numrange{0.06}{0.25}$), suggesting that it operates in a qualitatively distinct regime and succeeds on a different subset of motifs (Table~\ref{app:tab:motifbench}). Across BILBY-PRO, RFDiffusion and Proteina again achieve the broadest target coverage, indicating strong target-level reliability. ESM3 solves fewer targets but achieves comparatively high MotifBench scores on those it does solve. FrameFlow and Genie-2 remain intermediate, while DPLM-2 performs weak under both metrics.

Correlations broadly match the trends from the original 30-case MotifBench set, but are lower overall on the more diverse BILBY-PRO benchmark. RFDiffusion, Proteina, FrameFlow, and Genie2 remain substantially correlated ($r \approx \numrange{0.53}{0.79}$), indicating similar notions of motif difficulty, whereas ESM3 again shows weak correlations with all other methods ($r \approx \numrange{0.11}{0.26}$), reinforcing that it operates in a distinct regime from the diffusion-based scaffold generators.

Stratification by broad functional class shows that benchmark difficulty is motif-dependent. Nucleic-acid-binding/transcription and membrane/transport cases are comparatively tractable, whereas metal/co-factor-binding and secreted defense/toxin/inhibitor cases are more challenging, although category-level conclusions remain limited by small sample sizes. Overall, BILBY-PRO provides substantially greater resolution than the small curated reference set, separating broad target-level reliability from high-yield behavior on solved targets and revealing class-dependent trends not visible from aggregate coverage alone. These static results motivate the dynamics-aware analysis below. Coverage and MotifBench score measure recovery of static motif geometry and unique successful scaffolds under a fixed sampling budget, but not whether motifs remain stable in a physical ensemble. We therefore next test whether statically solved designs preserve motif geometry, scaffold structure, local contacts, and motif-dihedral landscapes under MD.

\section{BILBY-MD: Dynamics-Aware Evaluation}

To assess whether designs classified as successful by static motif-scaffolding evaluation exhibit dynamically credible behavior, we introduce BILBY-MD, an explicit-solvent all-atom MD evaluation framework. Conventional static evaluation relies on predicted conformations obtained through an inverse-folding and refolding pipeline, rather than directly assessing the conformational ensemble of the designed protein. In the standard MotifBench protocol adopted here, ProteinMPNN-designed sequences are refolded using ESMFold, which predicts structures from individual sequences without requiring multiple-sequence alignments (MSAs). Although ESMFold is explicitly designed for single-sequence prediction, the resulting success labels remain dependent on the accuracy and biases of the learned structure predictor. A design may therefore satisfy static motif-recovery criteria while exhibiting substantial scaffold drift, loss of motif-local contacts, or relaxation into a conformational basin that differs from the experimental reference. BILBY-MD provides a complementary physical evaluation by comparing designed proteins against simulations initiated from experimentally determined reference structures. Our primary analysis focuses on statically successful designs, supplemented by an additional analysis of unsuccessful cases in Appendix~\ref{app:md_unsolved}.

BILBY-MD probes this question at three complementary levels. First, we monitor structural drift through backbone Root Mean Square Displacement (RMSD) of both the full scaffold and the motif residues after alignment to the initial frame. Second, we compare time-averaged motif contact maps between designs and reference systems using overlap-, correlation-, and absolute-difference-based metrics to assess preservation of local interaction patterns. Third, we evaluate the conformational ensemble more globally by projecting motif backbone dihedral trajectories into common low-dimensional collective-variable spaces and comparing the resulting free-energy surfaces. These observables distinguish local motif retention from broader scaffold fidelity and give insight whether a benchmark solve reflects only geometric recoverability or also dynamical stability. 

\subsection{Simulation Details}

For each statically solved benchmark problem of the MotifBench test cases, we selected the solved design with the highest TM~score for subsequent MD simulation. This choice is conservative because it biases the MD evaluation toward the candidate most similar to an experimentally observed fold and therefore most likely to remain stable. The designed proteins were simulated for \SI{200}{ns}, while the corresponding experimental reference scaffolds were simulated for \SI{1}{\micro\second} and used as the baseline ensembles for downstream comparisons. All simulations were performed under physiological-like explicit-solvent conditions at \SI{300}{\kelvin}, \SI{1}{bar}, and \SI{150}{mM} NaCl. The full protocol including system-preparation, equilibration, production, and analysis protocol is given in appendix, \ref{app:md_simulation_protocol}.

\subsection{MD-support criteria}

The BILBY-MD analysis of statically solved designs covers 55 model/problem pairs with matching base-motif MD summaries. We treat a solved pair as MD-supported if at least three of four checks are satisfied: motif RMSD drift within \SI{1}{\angstrom}, scaffold RMSD drift within \SI{2}{\angstrom}, contact preservation above the contact thresholds, and agreement with the reference free-energy surfaces in both Principal Component Analysis (PCA) and Time-Lagged Independent Component Analysis (TICA) projections~\cite{tica}. This binary screen is deliberately conservative and should be read as a post hoc dynamical validation of static MotifBench solves rather than as a replacement for the original benchmark definition. Overall, 31/55 solved pairs (\SI{56.4}{\percent}) pass the MD-supported threshold, but only 8/55 (\SI{14.5}{\percent}) satisfy all four checks. The main message is therefore not that the static benchmark is uninformative: solved cases are better than unsolved cases for contact preservation and free-energy similarity. Rather, the static solved label is much more reliable for local motif correctness than for whole-scaffold dynamical stability. We give the full threshold rationale and sensitivity checks in appendix, \ref{app:md_support_heuristic}. Motif RMSD and contact-map similarity measure local motif retention. Scaffold RMSD measures global scaffold stability. Free-energy surface (FES) similarity measures whether the motif samples a conformational ensemble consistent with the reference rather than merely matching a representative structure. We use this local/global/ensemble framing throughout the MD results below.

\subsection{Model comparison}
\subsubsection{RMSD}

\begin{figure}[t]
    \centering
    \includegraphics[width=0.8\linewidth]{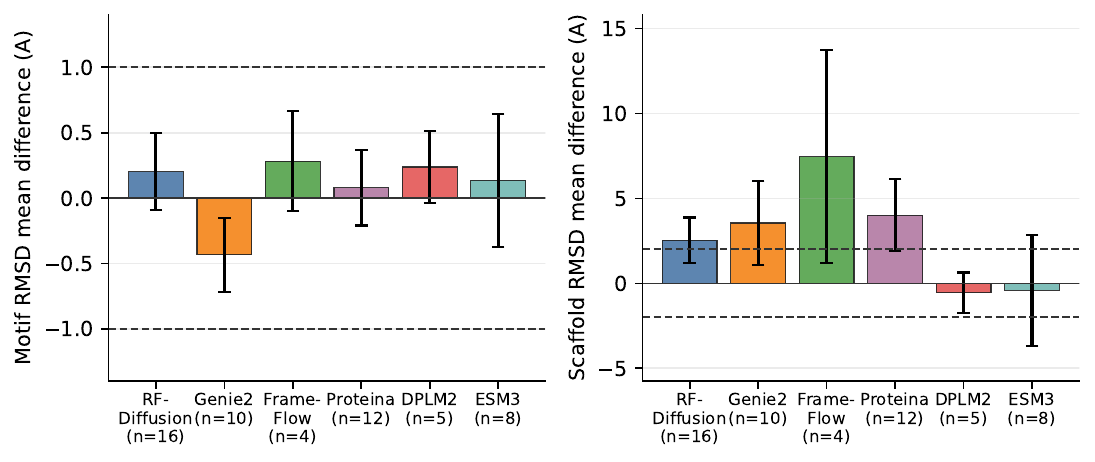}
    \caption{{\bf Model-level RMSD behavior for statically solved base-motif cases.} Values are averaged over solved pairs for each model; labels report the number of contributing solved pairs. The left panel shows motif RMSD mean difference relative to the reference, with dashed lines at $\pm \SI{1}{\angstrom}$. The right panel shows scaffold RMSD mean difference relative to the reference, with dashed lines at $\pm \SI{2}{\angstrom}$. Takeaway: motif RMSD remains near threshold for all models, while scaffold RMSD exposes substantial global drift for several methods.}
    \label{fig:md_rmsd_models}
\end{figure}

RMSD exposes the clearest separation between local motif retention and broader dynamical stability. Across solved pairs, 41/55 (\SI{74.5}{\percent}) pass the motif-drift check, whereas only 35/55 (\SI{63.6}{\percent}) pass the scaffold-drift check. Figure~\ref{fig:md_rmsd_models} summarizes motif and scaffold RMSD mean differences by model, averaging over the solved pairs available for each model and reporting the number of contributing pairs in the x-axis labels. The left panel shows motif RMSD mean difference relative to the reference, with dashed lines marking the $\pm \SI{1}{\angstrom}$ motif-drift threshold. The right panel shows the corresponding scaffold RMSD mean difference, with dashed lines marking the $\pm \SI{2}{\angstrom}$ scaffold-drift threshold.

At the model level, mean motif RMSD differences are within the \SI{1}{\angstrom} threshold for all six methods, indicating that the solved labels generally correspond to local motif retention. Scaffold RMSD is much more discriminating. FrameFlow shows the largest mean scaffold drift (\SI{7.46}{\angstrom}), followed by Proteina (\SI{4.01}{\angstrom}), Genie2 (\SI{3.56}{\angstrom}), and RFDiffusion (\SI{2.54}{\angstrom}), all above the scaffold threshold. DPLM2 and ESM3 have small signed mean scaffold differences (\num{-0.54} and \SI{-0.40}{\angstrom}), indicating the design of more stable scaffold compared to the reference structure; notably, these are the discrete tokenized models that jointly generate sequence and backbone, suggesting that this representation may favor global scaffold consistency. Overall, RMSD indicates that static solves are usually locally motif-correct but not necessarily scaffold-stable.

\subsubsection{Contacts}

\begin{figure}[h!]
    \centering
    \includegraphics[width=0.8\linewidth]{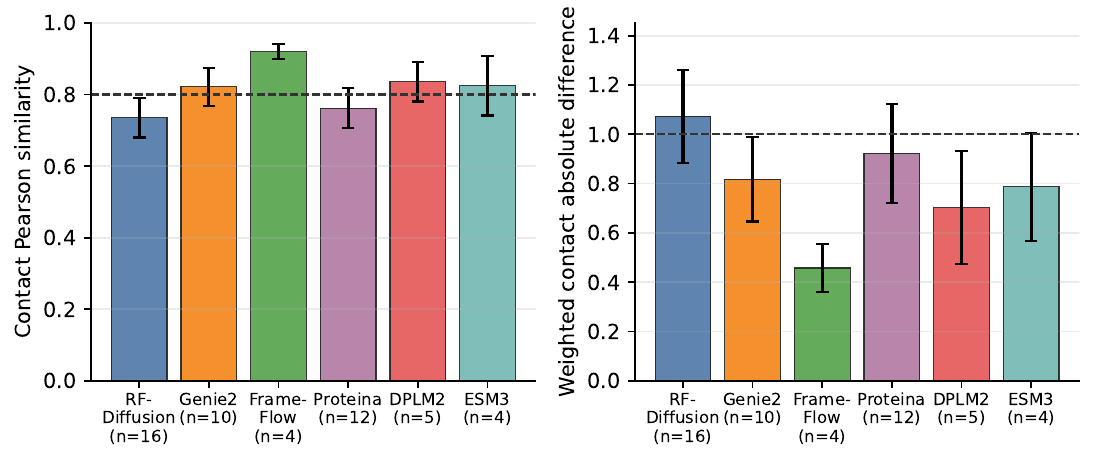}
    \caption{{\bf Model-level motif-contact preservation for statically solved cases.} The left panel shows contact Pearson similarity; the right panel shows weighted contact-map absolute difference. Dashed lines mark the heuristic MD-support thresholds used in the analysis: Pearson similarity above \num{0.8} and weighted absolute difference below \num{1.0}. Takeaway: contact preservation is generally stronger than scaffold stability, especially for the smaller FrameFlow and DPLM2 solved sets.}
    \label{fig:md_contacts_models}
\end{figure}

Contact-based observables provide the strongest support for the original solved labels. Solved motifs have an overall high contact Pearson similarity and low weighted contact-map absolute difference. The weighted absolute-difference metric is a reference-normalized $L_1$ error between time-averaged motif contact maps; its full definition is given in section \ref{app:contact_absdiff}. Only \SI{10.5}{\percent} of unsolved rows exceed the median solved row in contact Pearson, and only \SI{31.6}{\percent} improve on the median solved row in weighted contact-map absolute difference. Figure~\ref{fig:md_contacts_models} shows these two contact metrics by model.

The contact panels are more discriminating than RMSD. FrameFlow has the strongest mean contact preservation among its four solved pairs, with Pearson similarity \num{0.92} and weighted absolute difference \num{0.46}. DPLM2, ESM3, and Genie2 also exceed the Pearson threshold on average (\numlist{0.84;0.83;0.82}) and remain below the absolute-difference threshold. Proteina is intermediate: its weighted absolute difference is still below \num{1.0}, but its Pearson similarity is lower (\num{0.76}). RFDiffusion has the broadest solved set, but the weakest mean contact profile among the high-coverage models, with Pearson similarity \num{0.74} and weighted absolute difference \num{1.07}. This suggests that RFDiffusion's breadth comes with a larger fraction of locally imperfect solved cases, while FrameFlow's smaller solved set is contact-preserving but not necessarily globally stable.

\subsubsection{FES}

Free-energy-surface comparisons give a complementary ensemble-level view~\cite{Schffler2025}. A solved design can have low RMSD in representative frames but still sample a different conformational basin from the reference. We therefore compare PCA- and TICA-projected motif dihedral landscapes against the reference trajectories. Across solved pairs, 27/55 (\SI{49.1}{\percent}) pass the landscape-similarity check. Solved cases outperform unsolved cases: mean solved-minus-unsolved differences are positive for both PCA FES Pearson correlation (\num{0.317}) and TICA FES Pearson correlation (\num{0.279}). Figure~\ref{fig:md_fes_models} separates the four model-level FES observables: PCA Pearson, PCA absolute difference, TICA Pearson, and TICA absolute difference. The FES panels partly recapitulate the contact ranking but also reveal additional differences. For example Genie2, Proteina, and DPLM2 are closer to the landscape threshold in PCA and fall below it in TICA, indicating weaker agreement with slow motif-dihedral modes. The FES analysis therefore catches failures that contact maps alone can miss: a model can preserve many pairwise contacts while still sampling a reference-inconsistent dynamical basin.

\subsection{Problem-wise performance}

\begin{figure}[t]
    \centering
    \includegraphics[width=0.85\linewidth]{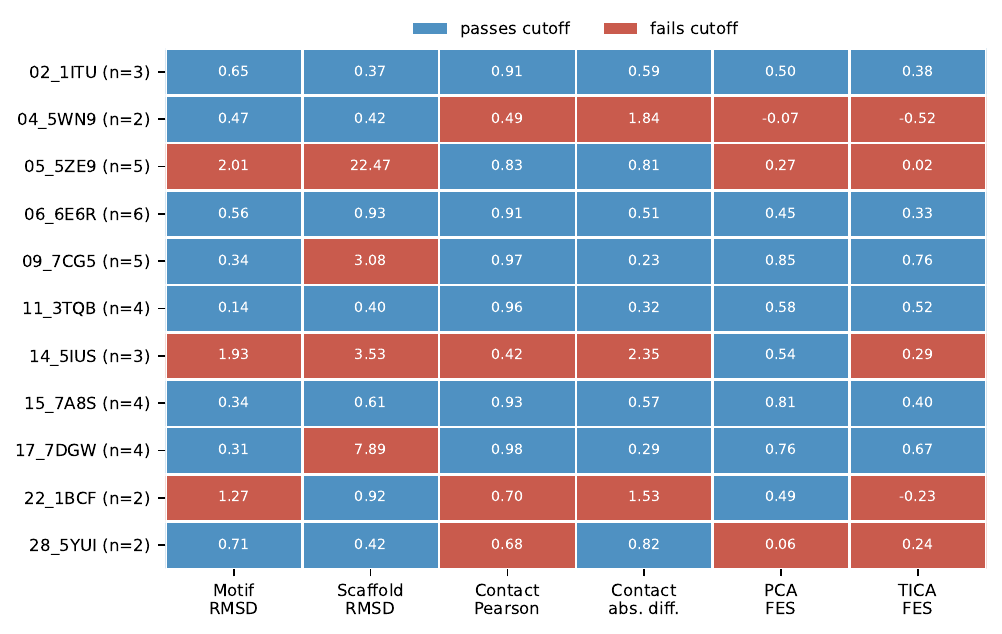}
    \caption{{\bf Raw MD metric profiles for representative MotifBench problems discussed in the text.} Rows are benchmark problems, with the number of contributing models solved by MotifBench in parentheses. Columns show motif RMSD mean difference, scaffold RMSD mean difference, contact Pearson similarity, weighted contact-map absolute difference, PCA FES Pearson similarity, and TICA FES Pearson similarity. Blue/red cells pass/fail the metric cutoffs as discussed in appendix~\ref{app:md_support_heuristic}. Takeaway: compact local motifs pass most checks, while interface-like or context-dependent cases fail in contacts, FES, or scaffold stability.}
    \label{fig:problem_threshold_heatmap}
\end{figure}

The MD behavior is strongly problem-dependent, and the problem-wise trends are more interpretable when connected to motif origin and biological role. Figure~\ref{fig:problem_threshold_heatmap} summarizes the representative cases discussed below using raw problem-level metric values averaged over solved models. Color indicates whether the raw value passes the corresponding MD-support cutoff. This makes the main biological pattern visually explicit: compact local motifs are mostly passing across columns, while interface-like or context-dependent motifs fail in contacts, FES, or scaffold RMSD.

The most robust cases are compact local motifs whose functional geometry can plausibly be stabilized in many scaffold contexts. The clearest example is \texttt{06\_6E6R}, the 75-residue ferredoxin [2Fe--2S] motif~\cite{Lovell2018}. It is solved by all six models and has an MD support rate of \num{1.0}, with mean scaffold drift of \SI{0.93}{\angstrom} and strong contact preservation (Pearson \num{0.91}). This is the most convincing case that a static MotifBench solve corresponds to a dynamically credible design. The biological interpretation is straightforward: ferredoxin function depends on a compact redox/co-factor geometry, and that local geometry appears scaffoldable without requiring a large missing interaction partner.

Two other favorable cases support the same conclusion. \texttt{11\_3TQB}, a parallel beta-strand/loop motif from dihydrofolate reductase~\cite{Franklin2011}, has an MD support rate of \num{0.75} and a strong-support rate of \num{0.50}, with very low scaffold drift (\SI{0.40}{\angstrom}) and high contact Pearson similarity (\num{0.96}). This case is less about a chemically elaborate active site and more about preserving a clean local topology, which appears to be dynamically well constrained. \texttt{15\_7A8S}, a \textit{de novo} TIM-barrel-derived design~\cite{Shanmugaratnam2020}, is similarly strong: support rate \num{0.75}, strong-support rate \num{0.50}, mean scaffold drift \SI{0.61}{\angstrom}, and contact Pearson \num{0.93}. Together, these cases show that both natural and synthetic motifs can survive MD when the motif is compact and structurally self-contained.

Several cases are MD-supported but require qualification. \texttt{09\_7CG5}, the anti-sigma factor RsgI4 sensor-domain case from the low-homology/orphan subset~\cite{Dong2020}, has support rate \num{1.0} and excellent contact and landscape agreement (contact Pearson \num{0.97}, PCA FES correlation \num{0.85}, TICA FES correlation \num{0.76}). However, its mean scaffold drift is \SI{3.08}{\angstrom}, above the scaffold threshold. \texttt{17\_7DGW}, a \textit{de novo} designed H4A2S fold~\cite{Xu2021}, is an even stronger example of this split: contact Pearson similarity is excellent (\num{0.98}) and FES agreement is high, but mean scaffold drift reaches \SI{7.89}{\angstrom}. These are not failed motifs in the local sense; rather, they show that motif-local dynamical fidelity can coexist with large-scale scaffold relaxation. This distinction matters for design applications: such scaffolds may be useful if the functional requirement is local presentation, but they should not be described as globally stable without further validation.

The clearest static-to-MD mismatch is \texttt{05\_5ZE9}, the V-ATPase P-loop/nucleotide-binding motif~\cite{Maruyama2019}. It is solved by five models but has an MD support rate of only \num{0.20}, driven by severe scaffold drift and weak TICA agreement. Interface and epitope problems are also weak: \texttt{04\_5WN9} and \texttt{14\_5IUS} both have zero MD-supported solved pairs, consistent with the idea that these motifs are slices of larger binding surfaces whose native stability depends on omitted partner or quaternary context~\cite{Fedechkin2017,Pascolutti2016}. The remaining enzyme/co-factor cases are mixed: \texttt{02\_1ITU} is reasonably convincing, whereas \texttt{22\_1BCF} and \texttt{28\_5YUI} fail enough contact or FES checks to lose MD support. Additional problem-level details are deferred to the appendix, \ref{app:md-eval}.

\subsection{Summary of MD evaluation}

The apparent discrepancy between preserved static rankings and incomplete MD support is not contradictory. Static metrics identify designs that can recover motif geometry after refolding, whereas MD reveals whether that geometry is dynamically maintained in a physically simulated ensemble. The MD evaluation therefore supports a narrower interpretation of MotifBench success than the static benchmark label alone might suggest. Static solves are enriched for designs that preserve motif contacts and motif-local free-energy landscapes, and this enrichment is visible both at the model level and relative to unsolved cases. However, the same label is a weaker indicator of whole-scaffold stability. RFDiffusion has the best overall balance of coverage and MD support, FrameFlow shows the strongest local-contact behavior in a small solved set but large scaffold drift, DPLM2 appears conservative with favorable scaffold RMSD but few solves, Genie2 and Proteina occupy the middle, and ESM3 remains difficult to assess because of missing FES projections.

Several limitations qualify this interpretation. The MD thresholds are heuristic rather than field-standard binary success definitions, although we provide a justification in the appendix, \ref{app:md_support_heuristic}. Our primary BILBY-MD analysis comprises 55 statically successful model--problem pairs. As an additional control, we evaluated 79 statically unsuccessful model--problem pairs spanning 28 original MotifBench problems (Appendix~\ref{app:md_unsolved}). Only 6/79 unsuccessful pairs satisfy the three-of-four MD-support criterion, compared with 31/55 successful pairs, and none satisfy all four checks. The \SI{200}{\nano\second} design trajectories may miss slower instabilities, while the reference trajectories are longer (\SI{1}{\micro\second}), making the comparison asymmetric. Given its computational expense, BILBY-MD is intended as a complementary validation step for selected promising designs following inexpensive static screening, rather than for exhaustive evaluation of all generated structures. Finally, selecting the highest-TM-score candidate for each solved problem biases the MD evaluation toward comparatively stable candidates, whereas benchmarks typically reward novelty; observed instability is therefore a conservative signal rather than an artifact of selecting obviously weak designs.

The interpretation of static success is additionally limited by its dependence on the ProteinMPNN--ESMFold evaluation pipeline. Although ESMFold does not require explicit multiple-sequence alignments, its predictions may still be affected by model-specific biases and imperfect sequence--structure compatibility assessment. Moreover, improved structure-prediction accuracy does not necessarily imply improved discrimination between physically viable and nonviable designs; recent work has demonstrated that evolutionary information can bias refolding-based assessments toward native-like structures even for substantially altered sequences \cite{korbeld2026refolding}. Consequently, our results characterize the dynamical fidelity of designs selected under the established MotifBench evaluation protocol, rather than providing predictor-independent estimates of absolute scaffold-generation success.

BILBY-PRO is also shaped by limitations of the source database: PROSITE patterns are sequence-level signatures, are biased toward annotated and structurally resolved protein families, and tend to produce contiguous or near-contiguous motif definitions, limiting the representation of discontinuous multi-segment functional sites. Because the evaluated generative models were pretrained on different and partly undisclosed datasets, overlap with benchmark source structures or homologues cannot be excluded. BILBY-PRO should therefore not be interpreted as a strictly out-of-distribution evaluation.

Biologically, the most MD-favorable motifs are compact local geometries such as redox/co-factor sites and topology-defining motifs, whereas therapeutic epitopes and protein--protein interface fragments perform worst because they depend on broader structural context. Overall, static motif recovery should therefore be interpreted as evidence of local structural correctness rather than full dynamical stability of the scaffold.

\section{Conclusion}
Our results suggest that motif-scaffolding performance should not be treated as a single scalar quantity. BILBY-PRO separates target-level coverage from unique-solution yield: RFDiffusion and Proteina provide broad success rate across targets, whereas ESM3 is more selective but often higher-yield on the targets it solves. This distinction would be difficult to resolve from a small hand-curated benchmark alone.

BILBY-MD analysis further shows that static refolding success is informative but incomplete. Static solves are enriched for local motif recovery, but they do not guarantee global scaffold stability or reference-like conformational ensembles. Compact, self-contained motifs are more often MD-supported, whereas interface-like or context-dependent motifs fail more frequently. Thus, static recovery, unique-solution yield, and dynamical fidelity should be reported as complementary axes rather than collapsed into a single ranking. BILBY-MD evaluation therefore strengthens rather than replaces static benchmarking. Static foldability metrics are scalable and useful for screening, but they rely on learned structure predictors and primarily measure recoverability under a computational oracle. MD adds an orthogonal physical criterion: given a plausible starting point, does the design remain stable in an explicit-solvent ensemble? This criterion is not sufficient to establish biochemical function, but for motif scaffolding it should be regarded as a necessary downstream check.

\subsubsection*{Code and Data Availability}
BILBY-PRO~\cite{BILBY-PRO}, comprising 220 PROSITE-derived functional motif-scaffolding benchmark cases and the associated benchmark-construction code, is available at \url{https://huggingface.co/datasets/HFMI-PROFOUND/BILBY-PRO}. The BILBY-MD molecular dynamics evaluation and analysis framework is hosted at \url{https://github.com/PROFOUND-HFMI/BILBY-MD}. Both resources are developed within the PROFOUND project of the Helmholtz Foundation Model Initiative, whose Hugging Face organization is available at \url{https://huggingface.co/HFMI-PROFOUND}. The BILBY-MD results reported in this work evaluate designs from the original MotifBench benchmark, whereas BILBY-PRO is independently constructed and evaluated using the established static motif-scaffolding protocol.

\subsubsection*{Acknowledgments}
We gratefully acknowledge the Gauss Centre for Supercomputing e.V. and the John von Neumann Institute for Computing (NIC) for computing time on JUPITER at Jülich Supercomputing Centre, including resources awarded through the NIC Excellence project PROFOUND. 
E.d.B., M.S., S.R.S. and S.K. acknowledge support by the German Federal Ministry for Research, Technology and Space (BMFTR) under grant number 01IS24076B (FAIME). 
E.d.B., M.S., S.R.S., S.K. and A.S. acknowledge support from the Helmholtz Foundation Model Initiative (HFMI) through PROFOUND and Virtual Cell. E.d.B, M.S, J.S., S.R.S and A.S. gratefully acknowledge funding by the Helmholtz Association's Initiative and Networking Fund through the Helmholtz AI platform. 
A.Y. and C.H. were partially funded by the Center for Advanced Systems Understanding (CASUS), which is financed by Germany’s Federal Ministry of Research, Technology, and Space (BMFTR), and by the Saxon Ministry for Science, Culture, and Tourism (SMWK), with tax funds on the basis of the budget approved by the Saxon State Parliament. C.H. was also supported by the National Research Foundation of Korea (NRF) grant funded by the Korea government (Ministry of Science and ICT) (No. RS-2023-00254343).
The funding bodies acknowledged above were not involved in the study design, data collection, analysis, decision to publish, or preparation of the manuscript.

\newpage{}

\printbibliography
\end{refsection}

\appendix
\renewcommand{\thetable}{A.\arabic{table}}
\setcounter{table}{0}

\renewcommand{\thefigure}{A.\arabic{figure}}
\setcounter{figure}{0}

\begin{refsection}
\section{BILBY-PRO Construction and Static Evaluation}

\subsection{BILBY-PRO Benchmark Construction}

For each eligible PROSITE entry, we parse the sequence pattern, associated documentation, and linked PDB structures. The pattern is cleaned and tokenized into residue-specific positions and wildcard segments, from which we derive regular expressions for chain-level sequence matching. We retain entries whose total matched span lies between 12 and 100 residues and that have at least one linked experimental structure.

For each linked PDB structure, we enumerate protein chains, search for matches to the corresponding PROSITE pattern, and map accepted matches back to residue indices and atomic coordinates. Contiguous matched positions are merged into motif segments. Short internal wildcard stretches may be absorbed into the motif and marked as redesignable, whereas longer wildcard stretches are preserved as non-motif scaffold context. Each extracted structural instance therefore records the motif residues, redesignable positions, non-motif context, and residue coordinates needed for downstream structural comparison and scaffold-generation export.

A PROSITE sequence pattern does not necessarily define a conserved three-dimensional motif. We therefore retain candidate motifs only when they are structurally conserved across multiple experimental instances. In the current selection, each candidate motif must have at least five structural instances, and the median pairwise backbone RMSD across instances must fall below a conservation threshold.

We also record conformational heterogeneity among retained motifs. Foldseek-derived cluster ratios are used to distinguish motifs that are predominantly represented by a single structural state from motifs that recur in multiple structural modes. We refer to the latter as \emph{multi-state} motifs. Multi-state motifs are not treated as failures of motif definition; instead, they identify PROSITE-linked signatures whose functional residues are compatible with more than one recurring structural configuration. Such cases are potentially informative for both static scaffold-generation benchmarks and downstream dynamics-aware evaluation.

In the final step, we remove cases that duplicate the original MotifBench problems and annotate each retained case by assembly context. Assemblies are tagged as \texttt{core}, \texttt{ribosomal}, or \texttt{huge\_assembly} based on coarse structural metadata.

BILBY-PRO follows the data layout expected by standard motif-scaffolding workflows. For each retained case, we provide a motif-only PDB file, the corresponding source structure, motif and non-motif segment annotations, redesignable residue indices, and contig specifications for scaffold generation. Coordinates and residue labels are normalized during export so that the cases can be consumed directly by existing generation and refolding pipelines.

The scarcity of Group 2 and Group 3 cases (according to the MotifBench definition,Table \ref{tab:prosite_groups}) is an expected consequence of deriving motifs from PROSITE sequence patterns. Most PROSITE-linked motifs are encoded as contiguous or near-contiguous within-chain signatures, so multi-segment motifs occur mainly when long internal wildcard spans are preserved as non-motif scaffold context. We therefore interpret the group distribution as a property of the source representation rather than as a failure of benchmark construction.

\begin{table}[h!]\label{tab:prosite_groups}
\caption{MotifBench-style group composition of BILBY-PRO. Groups 1--3 follow the MotifBench definitions; Group 4 denotes multi-conformational motifs identified by structural clustering.}
\label{tab:group_stats_appendix}
\centering
\begin{tabular}{lr}
\toprule
Group & Count \\
\midrule
Group 1 & 113 \\
Group 2 & 9 \\
Group 3 & 2 \\
Group 4 (multi-conformational) & 96 \\
\bottomrule
\end{tabular}
\end{table}

The public BILBY-PRO release includes minor contig and identifier corrections documented in its changelog; the reported historical evaluations were not recomputed, and the corrected contig was not used in the reported results.

\subsection{Protein generative models setup \& Scaffold generation}
We evaluated two benchmark panels: the original 30 MotifBench problems obtained from the published MotifBench repository, and the 220 independently constructed BILBY-PRO problems available at \url{https://huggingface.co/datasets/HFMI-PROFOUND/BILBY-PRO} ~\cite{BILBY-PRO}. Scaffolds were generated using the public upstream implementations of the six evaluated models, with model-specific input conversions where necessary. The resulting structures were assessed using the standard MotifBench ProteinMPNN--ESMFold evaluation protocol.

\subsection{RFDiffusion}

RFDiffusion \cite{rfdiffusion} is a denoising diffusion probabilistic model (DDPM) that adapts and fine-tunes the RoseTTAFold \cite{rosettafold} structure prediction network for generative protein design tasks. The model operates on residue-level rigid-body backbone frames, parametrized by translational and rotational components, and is SE(3)-equivariant. During training, protein backbone structures are progressively corrupted with rotational and translational noise, and the model learns to reverse this process by iteratively denoising noisy backbone frames toward realistic protein geometries. RFDiffusion leverages the strong geometric inductive biases learned during RoseTTAFold pre-training and additionally employs self-conditioning, where predictions from previous denoising steps are recycled during generation to improve stability and sample quality. In the context of motif scaffolding, motif residue coordinates are treated as constraints and passed throughout the reverse diffusion trajectory while only the surrounding scaffold residues are denoised. This enables the generation of diverse protein scaffolds that preserve the exact geometry and relative orientation of functional motifs, including motifs that are discontinuous in sequence space. 

\subsection{FrameFlow}

FrameFlow \cite{frameflow} is an SE(3) flow-matching generative model for protein backbone generation that represents each residue as a rigid-body frame parametrized by translations and rotations in three-dimensional space. Unlike traditional denoising diffusion probabilistic models, FrameFlow learns a continuous vector field that transports a prior distribution over backbone frames toward the data distribution along a probability path, rather than learning a stochastic reverse diffusion process. The model employs an SE(3)-equivariant architecture to preserve rotational and translational consistency while modelling residue-level geometric relationships. During motif scaffolding, FrameFlow can be extended using two complementary conditioning strategies: motif amortization, in which the model is explicitly trained to condition on fixed motif residues, and motif guidance, which guides an unconditional model toward satisfying motif constraints during sampling. In both settings, motif residue frames remain fixed while scaffold residues are initialized from noise and iteratively transformed along the learned flow trajectory to produce a coherent supporting scaffold. By directly modelling the flow vector field and integrating an ordinary differential equation during sampling, FrameFlow achieves substantially improved sampling efficiency and higher scaffold diversity compared to earlier SE(3)-based diffusion approaches. The resulting residue frames are subsequently converted into backbone atomic coordinates and exported as candidate scaffold structures.

\subsection{Genie2}
Genie2 \cite{genie2} is a denoising diffusion probabilistic model (DDPM) that operates exclusively on the $\mathrm{C}_{\alpha}$ coordinates. Rather than modelling full backbone frames in atomic detail, the model learns the score function of the distribution over protein $\mathrm{C}_{\alpha}$ traces using a Transformer network with relative positional features, while not strictly enforcing SE(3) equivariance as e.g. FrameFlow or RFDiffusion do. During training Genie2 learns to both conditional and unconditional score functions by being exposed to stochastically masked residue positions. At inference the model denoises a fully randomised $\mathrm{C}_{\alpha}$ trace until it converges to a geometry that is consistent with the fixed motif endpoints. During this loop the motif coordinates are softly reanchored, in contrast to being fixed with hard constraints, at each reverse diffusion step. After converging, the model then outputs a candidate for backbone coordinates. 

\subsection{Proteina}

Proteina \cite{proteina} is a large-scale generative foundation model for protein backbones that employs a flow-matching objective rather than standard diffusion. Similar to Genie2, the model operates primarily on protein $\mathrm{C}_{\alpha}$ coordinates, but utilizes a highly scaled non-equivariant Transformer architecture, with models exceeding 400M parameters, that avoids computationally expensive components such as triangle attention. Instead of learning a stochastic reverse diffusion process, Proteina learns a continuous vector field that transports an initial Gaussian distribution over backbone coordinates toward the distribution of realistic protein structures. A key feature of Proteina is its hierarchical fold-class conditioning framework, which enables semantic control over protein topology and secondary structure composition using CATH classifications. During motif scaffolding, the model performs conditional backbone generation by keeping motif coordinates fixed while transforming randomly initialized scaffold coordinates along the learned flow trajectory to produce a supporting protein scaffold. At inference the generated $\mathrm{C}_{\alpha}$ coordinates are exported as candidate backbone structures for downstream sequence design and evaluation. 

\subsection{DPLM2}
Contrary to models that explicitly predict the backbone structure from continuous features (Genie2, FrameFlow, etc.), Diffusion Protein Language Model 2 (DPLM2, \cite{dplm2}) uses a combination of amino acid and structural tokens. These are generated using a VQ-VAE structure tokenizer and which compresses the residue level geometry into a discrete codebook. During scaffold generation the model thus predicts both the most probable amino acid and structure token at each site. During this process, the motif tokens are fixed at their initial values while scaffold tokens are initialized as masks. Contrary to coordinate based methods, DPLM2 outputs both a sequence and a backbone prediction. At the end of generation, the structure token sequence is passed back through the VQ-VAEdecoder to reconstruct the $\mathrm{C}_{\alpha}$ coordinates, which are written to PDB files together with the placement metadata required by the MotifBench evaluation harness.

\subsection{ESM3}
ESM3 \cite{esm3} is a multimodal generative masked language model that jointly reasons over discrete tokens representing protein sequence, structure, and function. Unlike coordinate-based diffusion or flow models, ESM3 represents protein geometry using discretized structure tokens produced by a VQ-VAE-based structure tokenizer and performs generation through iterative masked token prediction across multiple representation tracks. The model is trained to recover masked sequence, structure, and functional annotations simultaneously, enabling conditional generation from partially specified prompts. During scaffold generation, motif residues can be provided as fixed sequence and structural constraints while the intervening scaffold tokens are initialized as masks and iteratively sampled by the model. Contrary to SE(3)-equivariant coordinate-based methods such as RFDiffusion or FrameFlow, ESM3 operates over discrete token sequences rather than directly modelling continuous three-dimensional geometry. As a result, the model is naturally best suited to scenarios where the relative sequence placement of motifs is predefined. While disjoint multi-motif scaffolding can in principle be attempted by specifying linker lengths or searching over motif arrangements, ESM3 does not explicitly optimize flexible spatial placement of multiple motifs in continuous space in the same manner as geometric diffusion-based scaffold generation methods. After generation, the predicted structure tokens are decoded back into backbone coordinates and exported as candidate scaffold structures.


\begin{table}[th!]
\caption{MotifBench leaderboard for the 30 MotifBench cases.}
\label{tab:static_leaderboard_motif}
\centering
\begin{tabular}{lll}
\toprule
Place & Model & MotifBench Score \\
\midrule
1 & Proteina & 37.16 \\
2 & ESM3 & 34.92 \\
3 & RFDiffusion & 27.72\\
4 & Genie2 & 17.39 \\
5 & FrameFlow & 12.19 \\
6 & DPLM2 &  6.98 \\
\bottomrule
\end{tabular}
\end{table}

\begin{table}[t]
\caption{Static evaluation leaderboard for BILBY-PRO (220 cases), using the MotifBench score.}
\label{tab:static_leaderboard_prosite}
\centering
\begin{tabular}{llS[table-format=2.2]}
\toprule
Place & Model & {MotifBench Score} \\
\midrule
1 & RFDiffusion & 62.20\\
2 & Proteina & 60.30 \\
3 & FrameFlow & 45.94 \\
4 & Genie2 & 32.35\\
5 & ESM3 &  24.98 \\
6 & DPLM2 & 5.77 \\
\bottomrule
\end{tabular}
\end{table}



\begin{table}[h!]
\caption{Static evaluation results for original MotifBench Testcases}
\label{app:tab:motifbench}
\resizebox{\textwidth}{!}{%
\begin{tabular}{|l|c|c|c|c|c|c|c|}
\hline
\textbf{Testcase}                  & \multicolumn{1}{l|}{\textbf{Group}} & \multicolumn{1}{l|}{\textbf{RFDiffusion}} & \multicolumn{1}{l|}{\textbf{FrameFlow}} & \multicolumn{1}{l|}{\textbf{Genie-2}} & \multicolumn{1}{l|}{\textbf{Proteina}} & \multicolumn{1}{l|}{\textbf{DPLM-2}} & \multicolumn{1}{l|}{\textbf{ESM3}} \\ \hline
01\_1LDB                           & \cellcolor[HTML]{FFFFFF}1           & \cellcolor[HTML]{B6D7A8}4                 & \cellcolor[HTML]{EA9999}0               & \cellcolor[HTML]{EA9999}0             & \cellcolor[HTML]{EA9999}0              & \cellcolor[HTML]{EA9999}0            & \cellcolor[HTML]{EA9999}0          \\ \hline
02\_1ITU                           & \cellcolor[HTML]{FFFFFF}1           & \cellcolor[HTML]{B6D7A8}4                 & \cellcolor[HTML]{EA9999}0               & \cellcolor[HTML]{B6D7A8}1             & \cellcolor[HTML]{EA9999}0              & \cellcolor[HTML]{EA9999}0            & \cellcolor[HTML]{B6D7A8}72         \\ \hline
03\_2CGA                           & \cellcolor[HTML]{FFFFFF}1           & \cellcolor[HTML]{EA9999}0                 & \cellcolor[HTML]{EA9999}0               & \cellcolor[HTML]{EA9999}0             & \cellcolor[HTML]{EA9999}0              & \cellcolor[HTML]{EA9999}0            & \cellcolor[HTML]{EA9999}0          \\ \hline
04\_5WN9                           & \cellcolor[HTML]{FFFFFF}1           & \cellcolor[HTML]{EA9999}0                 & \cellcolor[HTML]{EA9999}0               & \cellcolor[HTML]{B6D7A8}8             & \cellcolor[HTML]{B6D7A8}17             & \cellcolor[HTML]{EA9999}0            & \cellcolor[HTML]{EA9999}0          \\ \hline
05\_5ZE9                           & \cellcolor[HTML]{FFFFFF}1           & \cellcolor[HTML]{B6D7A8}31                & \cellcolor[HTML]{B6D7A8}54              & \cellcolor[HTML]{B6D7A8}32            & \cellcolor[HTML]{B6D7A8}33             & \cellcolor[HTML]{EA9999}0            & \cellcolor[HTML]{B6D7A8}39         \\ \hline
06\_6E6R                           & \cellcolor[HTML]{FFFFFF}1           & \cellcolor[HTML]{B6D7A8}56                & \cellcolor[HTML]{B6D7A8}60              & \cellcolor[HTML]{B6D7A8}54            & \cellcolor[HTML]{B6D7A8}92             & \cellcolor[HTML]{B6D7A8}84           & \cellcolor[HTML]{B6D7A8}19         \\ \hline
07\_6E6R                           & \cellcolor[HTML]{FFFFFF}1           & \cellcolor[HTML]{B6D7A8}74                & \cellcolor[HTML]{B6D7A8}34              & \cellcolor[HTML]{EA9999}0             & \cellcolor[HTML]{B6D7A8}90             & \cellcolor[HTML]{B6D7A8}10           & \cellcolor[HTML]{EA9999}0          \\ \hline
08\_7AD5                           & \cellcolor[HTML]{FFFFFF}1           & \cellcolor[HTML]{EA9999}0                 & \cellcolor[HTML]{EA9999}0               & \cellcolor[HTML]{B6D7A8}1             & \cellcolor[HTML]{EA9999}0              & \cellcolor[HTML]{EA9999}0            & \cellcolor[HTML]{EA9999}0          \\ \hline
09\_7CG5                           & \cellcolor[HTML]{FFFFFF}1           & \cellcolor[HTML]{B6D7A8}33                & \cellcolor[HTML]{B6D7A8}17              & \cellcolor[HTML]{B6D7A8}27            & \cellcolor[HTML]{B6D7A8}52             & \cellcolor[HTML]{B6D7A8}4            & \cellcolor[HTML]{EA9999}0          \\ \hline
10\_7WRK                           & \cellcolor[HTML]{FFFFFF}1           & \cellcolor[HTML]{B6D7A8}1                 & \cellcolor[HTML]{EA9999}0               & \cellcolor[HTML]{EA9999}0             & \cellcolor[HTML]{EA9999}0              & \cellcolor[HTML]{EA9999}0            & \cellcolor[HTML]{B6D7A8}1          \\ \hline
11\_3TQB                           & \cellcolor[HTML]{FFFFFF}2           & \cellcolor[HTML]{B6D7A8}6                 & \cellcolor[HTML]{EA9999}0               & \cellcolor[HTML]{B6D7A8}4             & \cellcolor[HTML]{B6D7A8}6              & \cellcolor[HTML]{EA9999}0            & \cellcolor[HTML]{B6D7A8}57         \\ \hline
12\_4JHW                           & \cellcolor[HTML]{FFFFFF}2           & \cellcolor[HTML]{EA9999}0                 & \cellcolor[HTML]{EA9999}0               & \cellcolor[HTML]{EA9999}0             & \cellcolor[HTML]{EA9999}0              & \cellcolor[HTML]{EA9999}0            & \cellcolor[HTML]{EA9999}0          \\ \hline
13\_4JHW                           & \cellcolor[HTML]{FFFFFF}2           & \cellcolor[HTML]{EA9999}0                 & \cellcolor[HTML]{EA9999}0               & \cellcolor[HTML]{EA9999}0             & \cellcolor[HTML]{EA9999}0              & \cellcolor[HTML]{EA9999}0            & \cellcolor[HTML]{EA9999}0          \\ \hline
14\_5IUS                           & \cellcolor[HTML]{FFFFFF}2           & \cellcolor[HTML]{B6D7A8}3                 & \cellcolor[HTML]{EA9999}0               & \cellcolor[HTML]{EA9999}0             & \cellcolor[HTML]{B6D7A8}6              & \cellcolor[HTML]{EA9999}0            & \cellcolor[HTML]{B6D7A8}46         \\ \hline
15\_7A8S                           & \cellcolor[HTML]{FFFFFF}2           & \cellcolor[HTML]{B6D7A8}3                 & \cellcolor[HTML]{EA9999}0               & \cellcolor[HTML]{B6D7A8}1             & \cellcolor[HTML]{B6D7A8}2              & \cellcolor[HTML]{EA9999}0            & \cellcolor[HTML]{B6D7A8}1          \\ \hline
16\_7BNY                           & \cellcolor[HTML]{FFFFFF}2           & \cellcolor[HTML]{B6D7A8}1                 & \cellcolor[HTML]{EA9999}0               & \cellcolor[HTML]{B6D7A8}2             & \cellcolor[HTML]{B6D7A8}2              & \cellcolor[HTML]{EA9999}0            & \cellcolor[HTML]{EA9999}0          \\ \hline
17\_7DGW                           & \cellcolor[HTML]{FFFFFF}2           & \cellcolor[HTML]{B6D7A8}4                 & \cellcolor[HTML]{EA9999}0               & \cellcolor[HTML]{B6D7A8}1             & \cellcolor[HTML]{B6D7A8}4              & \cellcolor[HTML]{EA9999}0            & \cellcolor[HTML]{B6D7A8}8          \\ \hline
18\_7MQQ                           & \cellcolor[HTML]{FFFFFF}2           & \cellcolor[HTML]{B6D7A8}1                 & \cellcolor[HTML]{EA9999}0               & \cellcolor[HTML]{EA9999}0             & \cellcolor[HTML]{EA9999}0              & \cellcolor[HTML]{EA9999}0            & \cellcolor[HTML]{EA9999}0          \\ \hline
19\_7MQQ                           & \cellcolor[HTML]{FFFFFF}2           & \cellcolor[HTML]{B6D7A8}10                & \cellcolor[HTML]{EA9999}0               & \cellcolor[HTML]{EA9999}0             & \cellcolor[HTML]{EA9999}0              & \cellcolor[HTML]{EA9999}0            & \cellcolor[HTML]{EA9999}0          \\ \hline
20\_7UWL                           & \cellcolor[HTML]{FFFFFF}2           & \cellcolor[HTML]{EA9999}0                 & \cellcolor[HTML]{EA9999}0               & \cellcolor[HTML]{EA9999}0             & \cellcolor[HTML]{EA9999}0              & \cellcolor[HTML]{EA9999}0            & \cellcolor[HTML]{EA9999}0          \\ \hline
21\_1B73                           & \cellcolor[HTML]{FFFFFF}3           & \cellcolor[HTML]{EA9999}0                 & \cellcolor[HTML]{EA9999}0               & \cellcolor[HTML]{EA9999}0             & \cellcolor[HTML]{EA9999}0              & \cellcolor[HTML]{EA9999}0            & \cellcolor[HTML]{EA9999}0          \\ \hline
22\_1BCF                           & \cellcolor[HTML]{FFFFFF}3           & \cellcolor[HTML]{B6D7A8}1                 & \cellcolor[HTML]{EA9999}0               & \cellcolor[HTML]{EA9999}0             & \cellcolor[HTML]{B6D7A8}1              & \cellcolor[HTML]{B6D7A8}1            & \cellcolor[HTML]{EA9999}0          \\ \hline
23\_1MPY                           & \cellcolor[HTML]{FFFFFF}3           & \cellcolor[HTML]{EA9999}0                 & \cellcolor[HTML]{EA9999}0               & \cellcolor[HTML]{EA9999}0             & \cellcolor[HTML]{EA9999}0              & \cellcolor[HTML]{EA9999}0            & \cellcolor[HTML]{EA9999}0          \\ \hline
24\_1QY3                           & \cellcolor[HTML]{FFFFFF}3           & \cellcolor[HTML]{EA9999}0                 & \cellcolor[HTML]{EA9999}0               & \cellcolor[HTML]{EA9999}0             & \cellcolor[HTML]{B6D7A8}4              & \cellcolor[HTML]{EA9999}0            & \cellcolor[HTML]{EA9999}0          \\ \hline
25\_2RKX                           & \cellcolor[HTML]{FFFFFF}3           & \cellcolor[HTML]{EA9999}0                 & \cellcolor[HTML]{EA9999}0               & \cellcolor[HTML]{EA9999}0             & \cellcolor[HTML]{B6D7A8}10             & \cellcolor[HTML]{EA9999}0            & \cellcolor[HTML]{EA9999}0          \\ \hline
26\_3B5V                           & \cellcolor[HTML]{FFFFFF}3           & \cellcolor[HTML]{EA9999}0                 & \cellcolor[HTML]{EA9999}0               & \cellcolor[HTML]{EA9999}0             & \cellcolor[HTML]{EA9999}0              & \cellcolor[HTML]{EA9999}0            & \cellcolor[HTML]{EA9999}0          \\ \hline
27\_4XOJ                           & \cellcolor[HTML]{FFFFFF}3           & \cellcolor[HTML]{EA9999}0                 & \cellcolor[HTML]{EA9999}0               & \cellcolor[HTML]{EA9999}0             & \cellcolor[HTML]{B6D7A8}10             & \cellcolor[HTML]{EA9999}0            & \cellcolor[HTML]{EA9999}0          \\ \hline
28\_5YUI                           & \cellcolor[HTML]{FFFFFF}3           & \cellcolor[HTML]{B6D7A8}3                 & \cellcolor[HTML]{EA9999}0               & \cellcolor[HTML]{EA9999}0             & \cellcolor[HTML]{B6D7A8}8              & \cellcolor[HTML]{EA9999}0            & \cellcolor[HTML]{EA9999}0          \\ \hline
29\_6CPA                           & \cellcolor[HTML]{FFFFFF}3           & \cellcolor[HTML]{EA9999}0                 & \cellcolor[HTML]{EA9999}0               & \cellcolor[HTML]{EA9999}0             & \cellcolor[HTML]{EA9999}0              & \cellcolor[HTML]{B6D7A8}1            & \cellcolor[HTML]{EA9999}0          \\ \hline
30\_7UWL                           & \cellcolor[HTML]{FFFFFF}3           & \cellcolor[HTML]{EA9999}0                 & \cellcolor[HTML]{EA9999}0               & \cellcolor[HTML]{EA9999}0             & \cellcolor[HTML]{EA9999}0              & \cellcolor[HTML]{EA9999}0            & \cellcolor[HTML]{EA9999}0          \\ \hline
\textbf{\# of Solved Original: 30} & \multicolumn{1}{l|}{}               & \textbf{16}                               & \textbf{4}                              & \textbf{10}                           & \textbf{15}                            & \textbf{5}                           & \textbf{8}                         \\ \hline
\end{tabular}%
}           
\end{table}

\section{Static evaluation metrics}
\label{app:static_metrics}

Coverage and the MotifBench score quantify related but distinct aspects of scaffold-generation performance. Let $M$ denote the number of benchmark problems and let $U_i$ be the number of unique successful scaffold clusters obtained for problem $i$. Coverage measures the fraction of problems solved at least once,
\begin{equation}
    \mathrm{Coverage}
    = \frac{1}{M}\sum_{i=1}^{M}\mathbb{I}[U_i>0].
\end{equation}
It is therefore a target-level reliability metric: a problem contributes equally whether it has one successful solution or many.

The benchmark-specific MotifBench score instead rewards the number of unique successful solutions per problem,
\begin{equation}
    \mathrm{MotifBenchScore}
    = \frac{1}{M}\sum_{i=1}^{M}(100+\alpha)\frac{U_i}{\alpha+U_i},
    \qquad \alpha=5.
\end{equation}
The concave transformation gives diminishing returns as additional unique solutions are found for an already well-solved problem. Coverage and MotifBench score should therefore be interpreted together: the former measures how broadly a method solves the benchmark, whereas the latter additionally captures the yield of structurally distinct successful solutions under the fixed sampling budget.

\section{BILBY-MD simulation protocol}
\label{app:md_simulation_protocol}

The structures entering the MD evaluation are the all-atom structures produced by the MotifBench pipeline during the evaluation of each of the models under study. In that pipeline, each predicted scaffold is sequence-designed with ProteinMPNN, generating eight candidate sequences, and each sequence is subsequently refolded with ESMFold. A case is classified as solved if it passes the benchmark self-consistency criterion, namely an RMSD below \SI{2}{\angstrom} between the original predicted scaffold structure and the corresponding ESMFold refolding, together with the remaining benchmark filters. For the BILBY-MD analysis, we do not simulate every solved structure associated with a given benchmark problem. Instead, from all solved cases for a problem, we select the structure with the highest TM score. This choice is intentionally conservative: a higher TM score indicates greater similarity to a known structure in the PDB and therefore biases the MD study toward the candidate that should be most likely to remain stable under simulation. This differs from the novelty-oriented interpretation of the benchmark itself, where lower TM scores score higher because they reward more structurally novel scaffolds.

All molecular-dynamics simulations were carried out with GROMACS 2024.3 using an all-atom AMBER99SB-ILDN protein force field and explicit TIP3P water. Structures classified as solved by the static benchmark were converted to all-atom simulation systems, placed in dodecahedral simulation boxes with a minimum solute--box distance of \SI{3.0}{\nano\metre}, solvated, neutralized, and brought to an NaCl concentration of \SI{150}{mM}. In addition, for each benchmark problem we simulated the corresponding reference scaffold, i.e., the experimental PDB structure underlying the problem definition, using the same preparation and simulation protocol; these reference trajectories were run for \SI{1}{\micro\second} and served as the baseline ensemble for all downstream comparisons. After steepest-descent energy minimization, each system was equilibrated in two stages: a \SI{50}{\pico\second} NVT phase at \SI{300}{\kelvin} using a velocity-rescaling thermostat, followed by a \SI{50}{\pico\second} NPT phase at \SI{300}{\kelvin} and \SI{1}{bar} using isotropic Parrinello--Rahman pressure coupling. Production simulations were then performed in the NPT ensemble with a \SI{2}{\femto\second} time step, Nos\'e--Hoover temperature coupling at \SI{300}{\kelvin}, and isotropic Parrinello--Rahman pressure coupling at \SI{1}{bar}. Bonds involving hydrogen were constrained with LINCS, periodic boundary conditions were applied in all dimensions, and long-range electrostatics were treated with particle-mesh Ewald. During production, real-space Coulomb and van der Waals interactions were evaluated with \SI{1.2}{\nano\metre} cutoffs. For the solved designs analyzed here, production trajectories were propagated for \SI{200}{\nano\second}, whereas the reference scaffold simulations were propagated for \SI{1}{\micro\second} under otherwise identical conditions. For analysis, trajectories were centered and aligned to the initial frame. Scaffold and motif backbone RMSD traces were then computed, motif contact maps were averaged over the trajectory, and motif backbone $\phi/\psi$ dihedrals were projected into shared PCA and TICA spaces derived from the corresponding reference simulations in order to compare free-energy surfaces between designed and reference systems.

\subsection{BioEmu comparison on Proteina designs}
\label{app:bioemu_comparison}

We assessed BioEmu as an alternative using ProteinMPNN-designed sequences from statically solved Proteina cases. Across ten paired cases, BioEmu ensembles showed lower mean motif-contact overlap with the reference simulations than scaffold-initialized MD (Jaccard similarity: 0.49 versus 0.64); mean PCA-FES correlation also decreased from 0.33 to $-0.14$ across seven evaluable pairs. More fundamentally, sequence-conditioned ensemble generation does not directly test whether a particular designed scaffold maintains its motif geometry during structural relaxation, which is the objective of our evaluation. We therefore retained scaffold-initialized MD; replacing it with BioEmu would introduce ensemble-prediction uncertainty into the assessment of scaffold stability.

\section{BILBY-MD Support Criteria and Sensitivity}
\label{app:md_support_heuristic}

The MD-support label is a transparent post hoc screen applied to statically solved model/problem pairs. It is not intended to redefine the MotifBench task. We use four checks that correspond to the three physical observables analyzed in the main text. A pair passes the motif-RMSD check if the absolute motif RMSD mean difference from the reference is at most 1~\AA. It passes the scaffold-RMSD check if the absolute scaffold RMSD mean difference is at most 2~\AA. It passes the contact check if the motif contact-map Pearson similarity is at least 0.8 and the weighted contact-map absolute difference is at most 1.0. It passes the landscape check if both PCA and TICA FES Pearson correlations are at least 0.3. A pair is called MD-supported if it passes at least three of these four checks, and strongly MD-supported if it passes all four.

The 3/4 rule is deliberately intermediate. A 2/4 rule would count cases with only partial local agreement as supported, whereas a 4/4 rule would reject cases where one observable is missing or one mode of comparison is noisy despite otherwise consistent behavior. In the current base-motif analysis, 31/55 solved pairs pass the 3/4 rule and 8/55 pass all four checks. We therefore performed a sensitivity analysis over both the number of required checks and the numerical cutoffs. Table~\ref{tab:md_cutoff_sensitivity_settings} summarizes the strict, baseline, and lenient settings used for each check.

\begin{table}[h!]
    \centering
    \caption{{\bf Cutoff settings used for the MD-support sensitivity analysis.} Motif RMSD and scaffold RMSD are absolute mean differences from the reference trajectory. The contact check requires both the Pearson and absolute-difference criteria to pass. The FES check requires both PCA and TICA FES Pearson correlations to pass the listed threshold.}
    \label{tab:md_cutoff_sensitivity_settings}
    \begin{tabular}{lS@{}lS@{}lS@{}l}
        \toprule
        Check cutoff & {Strict} & & {Baseline} & & {Lenient} & \\
        \midrule
        Motif RMSD & \le 0.5 & \si{\angstrom} & \le 1.0 & \si{\angstrom} & \le 1.5 & \si{\angstrom} \\
        Scaffold RMSD & \le 1.0 & \si{\angstrom} & \le 2.0 & \si{\angstrom} & \le 3.0 & \si{\angstrom} \\
        Contact Pearson & \ge 0.9 & & \ge 0.8 & & \geq 0.7 & \\
        Contact absolute difference & \le 0.75 & & \le 1.0 & & \le 1.25 & \\
        PCA/TICA FES Pearson & \ge 0.5 & & \ge 0.3 & & \ge 0.2 & \\
        \bottomrule
    \end{tabular}
\end{table}

\begin{figure}[t]
    \centering
    \includegraphics[width=0.9\linewidth]{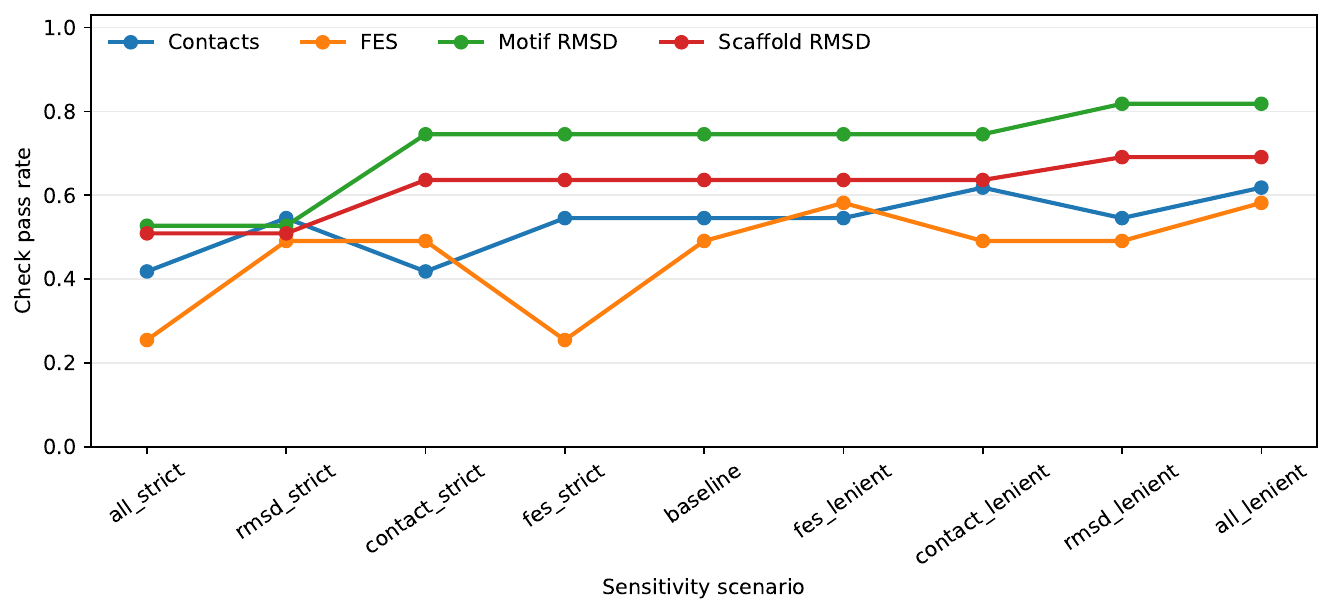}
    \caption{{\bf Sensitivity of individual MD-support checks to threshold choices.} Lines show the fraction of statically solved model/problem pairs passing each MD check under strict, baseline, and lenient cutoff settings, including isolated changes to RMSD, contact, and FES thresholds. Motif RMSD is consistently at least as permissive as scaffold RMSD and remains above it under the baseline and all-lenient settings, supporting the interpretation that static success is more predictive of motif-local geometry than of whole-scaffold dynamical stability.}
    \label{fig:md_cutoff_sensitivity_checks}
\end{figure}

Under the baseline thresholds, 43/55 statically solved model/problem pairs pass at least two checks, 31/55 pass at least three checks, and 8/55 pass all four checks. The absolute support rate changes as expected under stricter or more lenient thresholds: for the 3/4 MD-supported rule, support decreases to 18/55 under the all-strict setting and increases only modestly to 33/55 under the all-lenient setting. However, the qualitative interpretation is stable across the sensitivity settings. As shown in Figure~\ref{fig:md_cutoff_sensitivity_checks}, the motif-RMSD check passes more frequently than the scaffold-RMSD check at baseline (41/55 versus 35/55), under the all-strict setting (29/55 versus 28/55), and under the all-lenient setting (45/55 versus 38/55). Thus, static MotifBench success is more consistently associated with motif-local geometric recovery than with whole-scaffold dynamical stability. The contact and FES checks provide complementary local and ensemble-level information, but they are not uniformly less stringent than scaffold RMSD; for this reason, we interpret the sensitivity analysis as supporting a narrow conclusion about local motif recovery rather than a guarantee of full dynamical fidelity.

\subsection{Statically Unsolved Cases}
\label{app:md_unsolved}
As an additional validation of BILBY-MD, we evaluated 79 statically unsuccessful model--problem pairs spanning 28 original MotifBench problems. We applied the same four MD-support checks used for the statically successful designs. Table~\ref{tab:failed_case_overview} summarizes the results.

\begin{figure}[h!]
    \centering
    \includegraphics[width=0.8\linewidth]{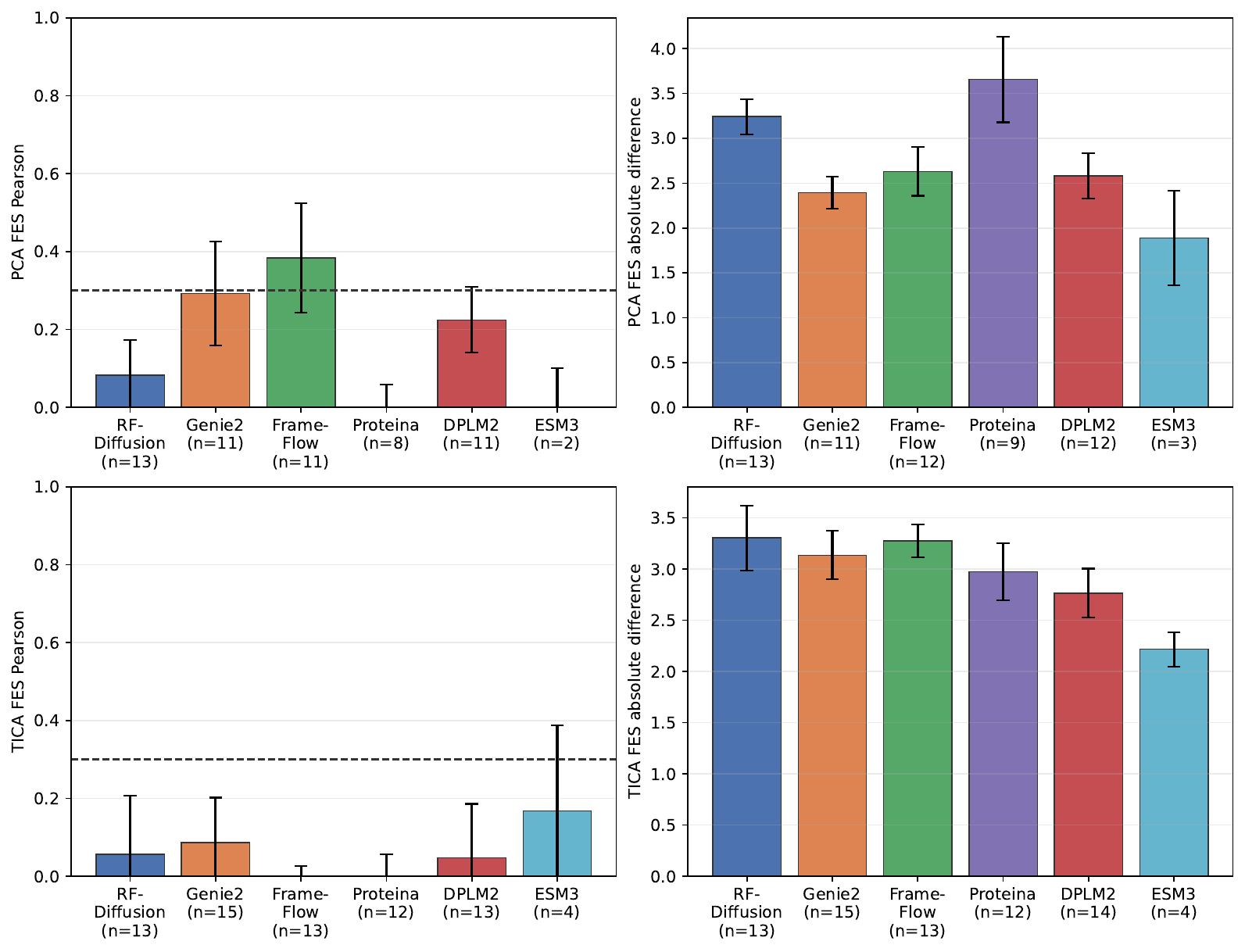}
    \caption{{\bf Model-level free-energy-surface similarity for statically unsolved base-motif cases.} Bars show model means and error bars show standard errors over unsolved pairs. Panels show PCA Pearson correlation, PCA absolute difference, TICA Pearson correlation, and TICA absolute difference between designed and reference motif-dihedral landscapes. Dashed lines in the Pearson panels mark the 0.3 threshold used in the MD sanity check. Takeaway: FES agreement is considerably worse than RMSD or scaffold, pointing to a clear dynamical failure of the statically unsolved simulations.}
    \label{fig:failed_fes}
\end{figure}

\begin{table}[h!]
\centering
\caption{BILBY-MD support outcomes for 79 statically unsuccessful model--problem pairs spanning 28 original MotifBench problems.}
\label{tab:failed_case_overview}
\begin{tabular}{lrl}
\toprule
MD-support result & \multicolumn{2}{r}{Negative-case model/problem pairs} \\
\midrule
Motif-RMSD check passes                  & 42/79 & (\SI{53.2}{\percent}) \\
Scaffold-RMSD check passes               & 53/79 & (\SI{67.1}{\percent}) \\
Combined contact check passes            & 9/79  & (\SI{11.4}{\percent}; 9/75 evaluable) \\
Combined PCA/TICA landscape check passes & 6/79  & (\SI{7.6 }{\percent}; 6/55 evaluable) \\
At least 2/4 checks pass                 & 37/79 & (\SI{46.8}{\percent}) \\
At least 3/4 checks pass                 & 6/79  & (\SI{7.6 }{\percent}) \\
All 4/4 checks pass                      & 0/79  & (\SI{0.0 }{\percent}) \\
\bottomrule
\end{tabular}
\end{table}
What stands out about the unsolved case simulations is that none pass 4/4 of the metrics. Additionally only \SI{7.6}{\percent} pass three out of four checks, which compared to the \SI{56}{\percent} for the statically solved cases, is a reassuring signal. Table~\ref{tab:failed_case_overview}. 
\begin{figure}[h!]
    \centering
    \includegraphics[width=0.8\linewidth]{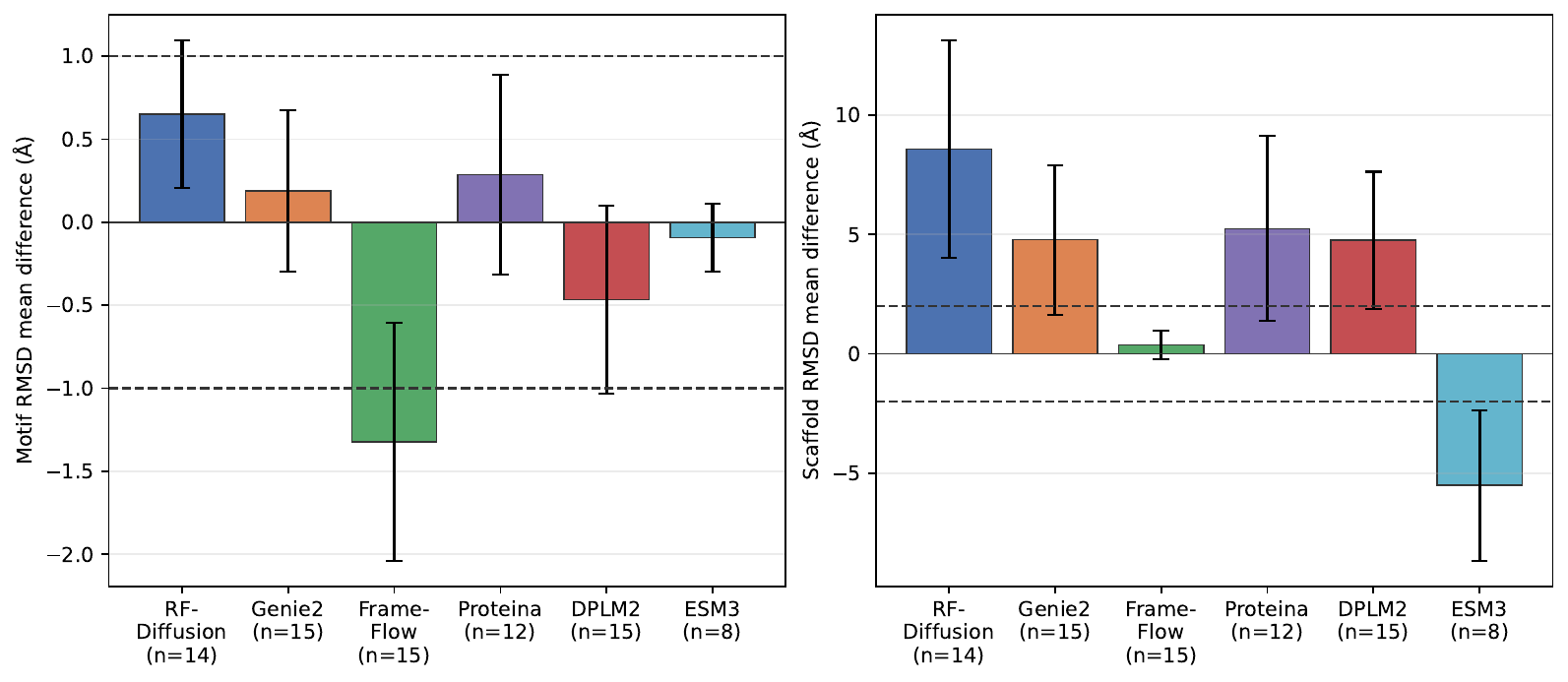}
    \caption{{\bf Model-level RMSD behavior for statically unsolved base-motif cases.} Values are averaged over unsolved pairs for each model; labels report the number of contributing unsolved pairs. The left panel shows motif RMSD mean difference relative to the reference, with dashed lines at $\pm 1$~\AA. The right panel shows scaffold RMSD mean difference relative to the reference, with dashed lines at $\pm 2$~\AA. Takeaway: similarily to the statically solved case evaluation the motif RMSD mean difference remains more conserved than the scaffold RMSD mean difference.}
    \label{fig:failed_rmsd}
\end{figure}
\begin{figure}[h!]
    \centering
    \includegraphics[width=0.8\linewidth]{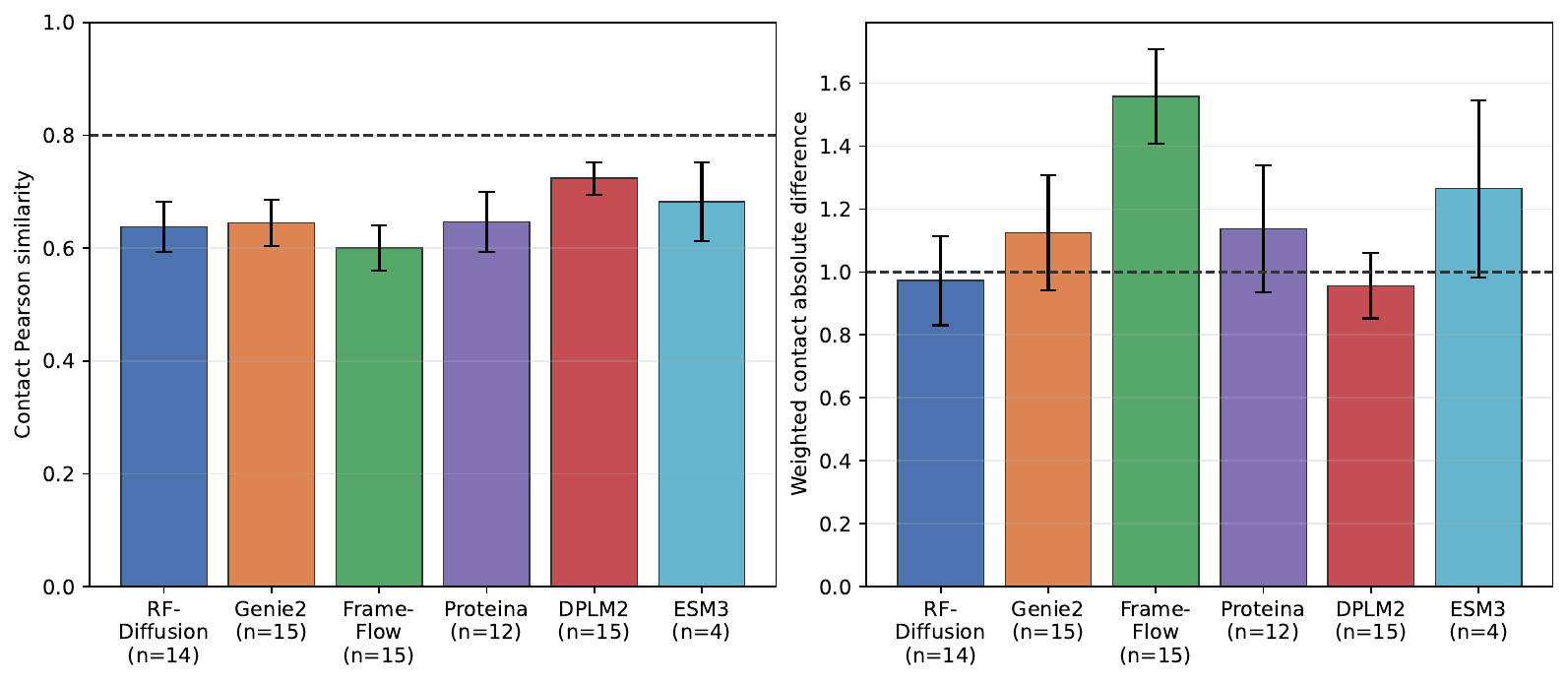}
    \caption{{\bf Model-level motif-contact preservation for statically unsolved cases.} The left panel shows contact Pearson similarity; the right panel shows weighted contact-map absolute difference. Dashed lines mark the heuristic MD-support thresholds used in the analysis: Pearson similarity above 0.8 and weighted absolute difference below 1.0. Takeaway: Especially when it comes to the contact Pearson score the models perform quite similarly, however none manage to breach the 0.8 threshold.}
    \label{fig:failed_contact}
\end{figure}
Specifically Figure~\ref{fig:failed_fes} paints a clear picture. 
Statically unsuccessful designs are substantially less likely to exhibit reference-consistent dynamical behavior, particularly in the free-energy-landscape comparisons. Nevertheless, 6 of the 79 unsuccessful pairs satisfy the three-of-four MD-support criterion, indicating that static failure does not categorically exclude dynamical support. 
\begin{figure}[h!]
    \centering
    \includegraphics[width=0.8\linewidth]{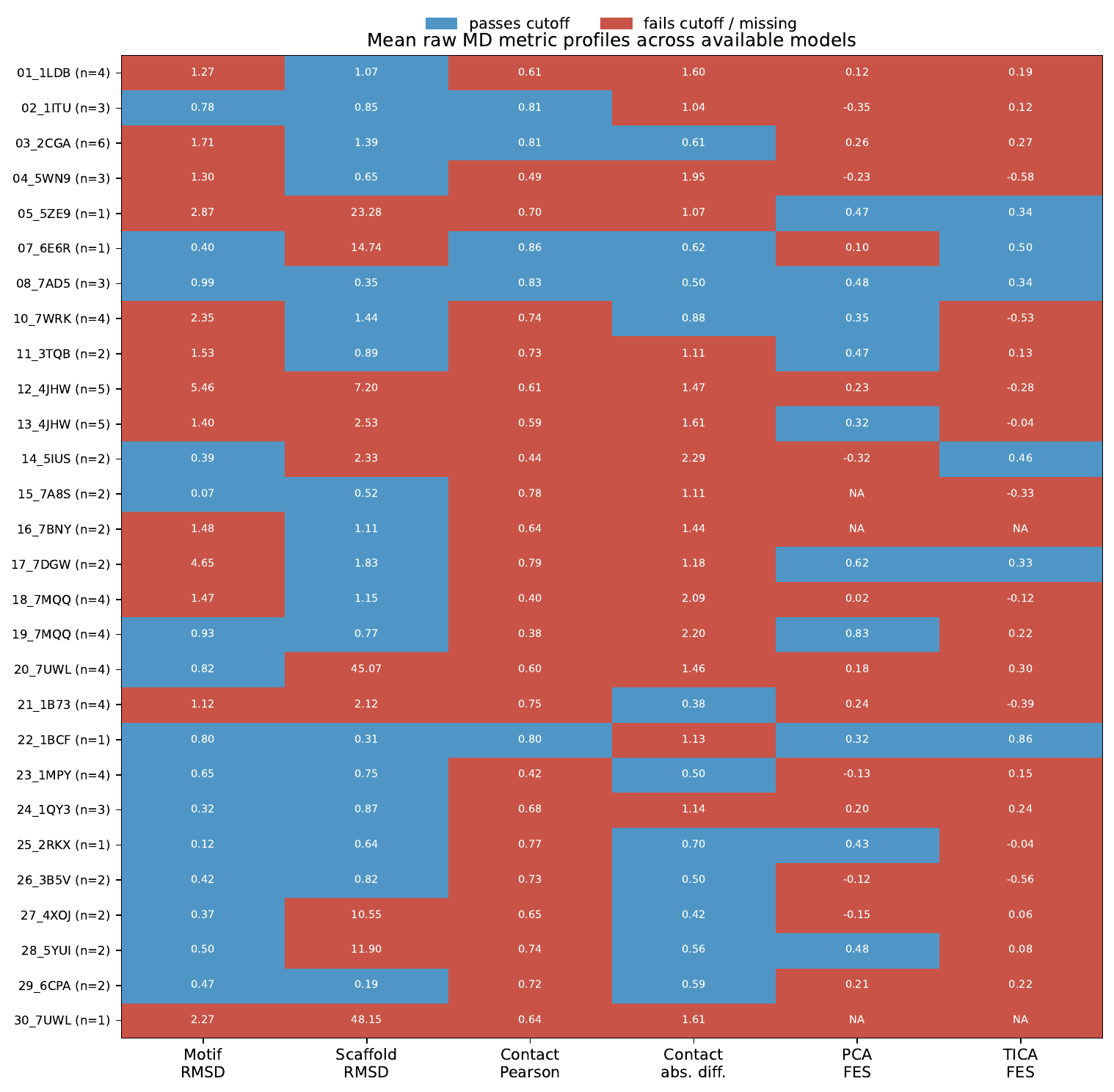}
    \caption{{\bf Raw MD metric profiles for representative MotifBench problems discussed in the text.} Rows are benchmark problems, with the number of contributing models not-solved by MotifBench in parentheses. Columns show motif RMSD mean difference, scaffold RMSD mean difference, contact Pearson similarity, weighted contact-map absolute difference, PCA FES Pearson similarity, and TICA FES Pearson similarity. Blue/red cells pass/fail the metric cutoffs as discussed in appendix~\ref{app:md_support_heuristic}. Takeaway: Much lower pass-rate than the statically solved cases, with no clear distinction between models.}
    \label{fig:failed_metric_profiles}
\end{figure}
\section{Contact-map absolute-difference metric}
\label{app:contact_absdiff}

For each simulated design and its corresponding reference system, we compute a time-averaged motif contact map. Each entry $C_{ij}$ summarizes how often, or how strongly, motif residues $i$ and $j$ are in contact over the trajectory. Pearson similarity compares the pattern of contact strengths after centering and scaling, and is therefore insensitive to some global changes in contact magnitude. We therefore also report an absolute-difference metric that measures the direct contact-map error.

Let $C^{\mathrm{design}}_{ij}$ and $C^{\mathrm{ref}}_{ij}$ denote the design and reference contact-map entries for motif-residue pair $(i,j)$. The unnormalized contact error is the L1 distance
\[
    D_{\mathrm{abs}} = \sum_{i<j} \left| C^{\mathrm{design}}_{ij} - C^{\mathrm{ref}}_{ij} \right|.
\]
The reported weighted absolute difference normalizes this error by the total reference contact weight,
\[
    D_{\mathrm{weighted}} =
    \frac{\sum_{i<j} \left| C^{\mathrm{design}}_{ij} - C^{\mathrm{ref}}_{ij} \right|}
         {\sum_{i<j} C^{\mathrm{ref}}_{ij}}.
\]
Thus, smaller values indicate better motif-contact preservation. A value near zero indicates nearly identical time-averaged contact maps, whereas a value near one means that the cumulative contact-map error is comparable to the total amount of reference contact signal. This metric complements Pearson similarity: Pearson asks whether strong and weak contacts occur in the same relative pattern, while the weighted absolute difference penalizes changes in the magnitude or loss of contacts directly.

\section{Relationship between PCA, TICA, and free-energy surfaces}
\label{app:fes_tica_pca}

PCA, TICA, and the free-energy-surface (FES) analysis are not three sequential or independent analyses. PCA and TICA define two alternative low-dimensional coordinate systems in which separate FES comparisons are performed. For each benchmark problem, motif-backbone $\phi/\psi$ dihedral trajectories from the reference simulation are used to define the corresponding principal-component (PC) and time-lagged independent-component (TIC) spaces. The reference trajectory and all corresponding designed trajectories are then projected into these same reference-derived spaces. This places the design and reference probability distributions in a common coordinate system and allows their free-energy landscapes to be compared directly.

The two projections emphasize different properties of the same MD trajectories. PCA identifies directions of largest structural variance and therefore describes the dominant geometrical spread of the motif ensemble. TICA instead identifies combinations with high time-lagged autocorrelation and is intended to emphasize slowly decorrelating conformational processes. For either projection, the sampled probability density $P(z_1,z_2)$ in the first two components is converted to a free-energy surface according to
\begin{equation}
    F(z_1,z_2) = -k_{\mathrm{B}}T\ln P(z_1,z_2) + C,
\end{equation}
where $C$ is an arbitrary additive constant. Design and reference surfaces are then compared using FES Pearson correlation and an absolute-difference measure. Thus, PCA-FES and TICA-FES provide complementary views of ensemble agreement rather than independent evidence from separate trajectories.

This distinction is also reflected in the MD-support screen. PCA and TICA do not count as two independent votes. Together they define a single landscape-agreement check, which is passed only when both the PCA-derived and TICA-derived FES Pearson correlations exceed the specified threshold (Appendix~\ref{app:md_support_heuristic}). The absolute-difference observables are reported as complementary descriptive measures but are not separate votes in the MD-support criterion.
\begin{figure}[h!]
    \centering
    \includegraphics[width=0.8\linewidth]{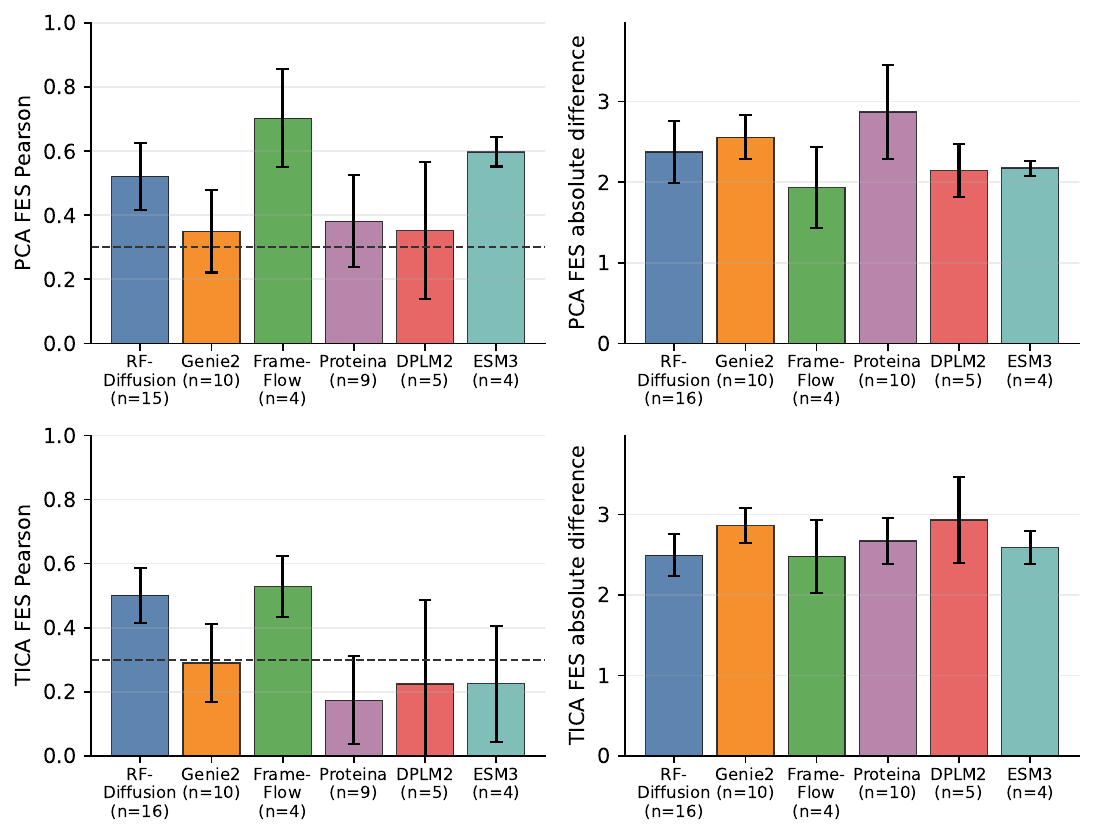}
    \caption{{\bf Model-level free-energy-surface similarity for solved base-motif cases.} Bars show model means and error bars show standard errors over solved pairs. Panels show PCA Pearson correlation, PCA absolute difference, TICA Pearson correlation, and TICA absolute difference between designed and reference motif-dihedral landscapes. Dashed lines in the Pearson panels mark the 0.3 threshold used in the MD sanity check. Takeaway: FES agreement separates local-contact preservation from ensemble-level agreement; many high-scoring models do not reproduce the ensemble dynamics even when they score high on the static benchmark.}
    \label{fig:md_fes_models}
\end{figure}

\section{MD-evaluation in depth}\label{app:md-eval}
Here we discuss the missing test cases of the indepth discussion of the main text.

The remaining active-site and co-factor cases are mixed rather than uniformly stable. \texttt{02\_1ITU}, the renal dipeptidase/cilastatin-associated motif, is reasonably convincing with support rate 0.67, low scaffold drift, and strong contact preservation. \texttt{29\_6CPA}, the carboxypeptidase A active-site case, is supported for its single solved pair. In contrast, \texttt{22\_1BCF} and \texttt{28\_5YUI}, both metal- or co-factor-associated natural enzyme motifs, have zero MD support in the solved-pair table, and \texttt{27\_4XOJ}, the trypsin catalytic-triad/oxyanion-hole case, has one unsupported solved pair with high scaffold drift. Thus, enzyme relevance alone does not guarantee MD robustness. What matters more is whether the motif is locally self-contained or depends on a larger fold context to enforce the native geometry.

At the provenance level, the low-homology/orphan subset is surprisingly heterogeneous rather than uniformly fragile. Cases such as \texttt{09\_7CG5}, \texttt{15\_7A8S}, and \texttt{17\_7DGW} are among the best locally preserved motifs, whereas plant-effector-style cases such as \texttt{08\_7AD5} and the \texttt{7MQQ} pair are weak. The design-paper-derived subset is also mixed, but its aggregate behavior is dominated by the severe \texttt{05\_5ZE9} failure. The older epitope/interface subset is the clearest negative category, with no MD-supported solved pairs among the analyzed \texttt{04\_5WN9} and \texttt{14\_5IUS} examples.

\printbibliography
\end{refsection}


\end{document}